\documentclass{article}

\catcode`!=11 
 
\def\PiC{P\kern-.12em\lower.5ex\hbox{I}\kern-.075emC}
\def\PiCTeX{\PiC\kern-.11em\TeX}

\def\!ifnextchar#1#2#3{%
  \let\!testchar=#1%
  \def\!first{#2}%
  \def\!second{#3}%
  \futurelet\!nextchar\!testnext}
\def\!testnext{%
  \ifx \!nextchar \!spacetoken 
    \let\!next=\!skipspacetestagain
  \else
    \ifx \!nextchar \!testchar
      \let\!next=\!first
    \else 
      \let\!next=\!second 
    \fi 
  \fi
  \!next}
\def\\{\!skipspacetestagain} 
  \expandafter\def\\ {\futurelet\!nextchar\!testnext} 
\def\\{\let\!spacetoken= } \\  

\def\!tfor#1:=#2\do#3{%
  \edef\!fortemp{#2}%
  \ifx\!fortemp\!empty 
    \else
    \!tforloop#2\!nil\!nil\!!#1{#3}%
  \fi}
\def\!tforloop#1#2\!!#3#4{%
  \def#3{#1}%
  \ifx #3\!nnil
    \let\!nextwhile=\!fornoop
  \else
    #4\relax
    \let\!nextwhile=\!tforloop
  \fi 
  \!nextwhile#2\!!#3{#4}}

\def\!etfor#1:=#2\do#3{%
  \def\!!tfor{\!tfor#1:=}%
  \edef\!!!tfor{#2}%
  \expandafter\!!tfor\!!!tfor\do{#3}}

\def\!cfor#1:=#2\do#3{%
  \edef\!fortemp{#2}%
  \ifx\!fortemp\!empty 
  \else
    \!cforloop#2,\!nil,\!nil\!!#1{#3}%
  \fi}
\def\!cforloop#1,#2\!!#3#4{%
  \def#3{#1}%
  \ifx #3\!nnil
    \let\!nextwhile=\!fornoop 
  \else
    #4\relax
    \let\!nextwhile=\!cforloop
  \fi
  \!nextwhile#2\!!#3{#4}}

\def\!ecfor#1:=#2\do#3{%
  \def\!!cfor{\!cfor#1:=}%
  \edef\!!!cfor{#2}%
  \expandafter\!!cfor\!!!cfor\do{#3}}

\def\!empty{}
\def\!nnil{\!nil}
\def\!fornoop#1\!!#2#3{}

\def\!ifempty#1#2#3{%
  \edef\!emptyarg{#1}%
  \ifx\!emptyarg\!empty
    #2%
  \else
    #3%
  \fi}
 
\def\!getnext#1\from#2{%
  \expandafter\!gnext#2\!#1#2}%
\def\!gnext\\#1#2\!#3#4{%
  \def#3{#1}%
  \def#4{#2\\{#1}}%
  \ignorespaces}

\def\!getnextvalueof#1\from#2{%
  \expandafter\!gnextv#2\!#1#2}%
\def\!gnextv\\#1#2\!#3#4{%
  #3=#1%
  \def#4{#2\\{#1}}%
  \ignorespaces}

\def\!copylist#1\to#2{%
  \expandafter\!!copylist#1\!#2}
\def\!!copylist#1\!#2{%
  \def#2{#1}\ignorespaces}

\def\!wlet#1=#2{%
  \let#1=#2 
  \wlog{\string#1=\string#2}}
 
\def\!listaddon#1#2{%
  \expandafter\!!listaddon#2\!{#1}#2}
\def\!!listaddon#1\!#2#3{%
  \def#3{#1\\#2}}
 
\def\!rightappend#1\withCS#2\to#3{\expandafter\!!rightappend#3\!#2{#1}#3}
\def\!!rightappend#1\!#2#3#4{\def#4{#1#2{#3}}}

\def\!leftappend#1\withCS#2\to#3{\expandafter\!!leftappend#3\!#2{#1}#3}
\def\!!leftappend#1\!#2#3#4{\def#4{#2{#3}#1}}

\def\!lop#1\to#2{\expandafter\!!lop#1\!#1#2}
\def\!!lop\\#1#2\!#3#4{\def#4{#1}\def#3{#2}}

\def\!loop#1\repeat{\def\!body{#1}\!iterate}
\def\!iterate{\!body\let\!next=\!iterate\else\let\!next=\relax\fi\!next}
 
\def\!!loop#1\repeat{\def\!!body{#1}\!!iterate}
\def\!!iterate{\!!body\let\!!next=\!!iterate\else\let\!!next=\relax\fi\!!next}
 
\def\!removept#1#2{\edef#2{\expandafter\!!removePT\the#1}}
{\catcode`p=12 \catcode`t=12 \gdef\!!removePT#1pt{#1}}

\def\placevalueinpts of <#1> in #2 {%
  \!removept{#1}{#2}}
 
\def\!mlap#1{\hbox to 0pt{\hss#1\hss}}
\def\!vmlap#1{\vbox to 0pt{\vss#1\vss}}
 
\def\!not#1{%
  #1\relax
    \!switchfalse
  \else
    \!switchtrue
  \fi
  \if!switch
  \ignorespaces}

\let\!!!wlog=\wlog              
\def\wlog#1{}    

\newskip\headingtoplotskip      
\newskip\linethickness         
\newskip\longticklength         
\newskip\plotsymbolspacing     
\newskip\shortticklength        
\newskip\stackleading           
\newskip\tickstovaluesleading   
\newskip\totalarclength        
\newskip\valuestolabelleading   

\newbox\!boxA                   
\newbox\!boxB                   
\newbox\!picbox                 
\newbox\!plotsymbol             
\newbox\!putobject              
\newbox\!shadesymbol            

\newcount\!countA               
\newcount\!countB               
\newcount\!countC               
\newcount\!countD               
\newcount\!countE               
\newcount\!countF               
\newcount\!countG               
\newcount\!fiftypt              
\newcount\!intervalno           
\newcount\!npoints              
\newcount\!nsegments            
\newcount\!ntemp                
\newcount\!parity               
\newcount\!scalefactor          
\newcount\!tickcase             

\newskip\!Xleft                
\newskip\!Xright               
\newskip\!Xsave                
\newskip\!Ybot                 
\newskip\!Ysave                
\newskip\!Ytop                 
\newskip\!angle                
\newskip\!arclength            
\newskip\!areabloc             
\newskip\!arealloc             
\newskip\!arearloc             
\newskip\!areatloc             
\newskip\!bshrinkage           
\newskip\!checkbot             
\newskip\!checkleft            
\newskip\!checkright           
\newskip\!checktop             
\newdimen\!dimenA               
\newdimen\!dimenB               
\newdimen\!dimenC               
\newdimen\!dimenD               
\newdimen\!dimenE               
\newdimen\!dimenF               
\newdimen\!dimenG               
\newdimen\!dimenH               
\newdimen\!dimenI               
\newdimen\!distacross           
\newdimen\!downlength           
\newdimen\!dp                   
\newdimen\!dshade               
\newdimen\!dxpos                
\newdimen\!dxprime              
\newdimen\!dypos                
\newdimen\!dyprime              
\newdimen\!ht                   
\newdimen\!leaderlength         
\newdimen\!lshrinkage           
\newdimen\!midarclength         
\newdimen\!offset               
\newdimen\!plotheadingoffset    
\newdimen\!plotsymbolxshift     
\newdimen\!plotsymbolyshift     
\newdimen\!plotxorigin          
\newdimen\!plotyorigin          
\newdimen\!rshrinkage           
\newdimen\!shadesymbolxshift    
\newdimen\!shadesymbolyshift    
\newdimen\!tshrinkage           
\newdimen\!uplength             
\newdimen\!wd                   
\newdimen\!xB                   
\newdimen\!xC                   
\newdimen\!xE                   
\newdimen\!xM                   
\newdimen\!xS                   
\newdimen\!xaxislength          
\newdimen\!xdiff                
\newdimen\!xleft                
\newdimen\!xloc                 
\newdimen\!xorigin              
\newdimen\!xpivot               
\newdimen\!xpos                 
\newdimen\!xprime               
\newdimen\!xright               
\newdimen\!xshade               
\newdimen\!xshift               
\newdimen\!xtemp                
\newdimen\!xunit                
\newdimen\!xxE                  
\newdimen\!xxM                  
\newdimen\!xxS                  
\newdimen\!xxloc                
\newdimen\!yB                   
\newdimen\!yC                   
\newdimen\!yE                   
\newdimen\!yM                   
\newdimen\!yS                   
\newdimen\!yaxislength          
\newdimen\!ybot                 
\newdimen\!ydiff                
\newdimen\!yloc                 
\newdimen\!yorigin              
\newdimen\!ypivot               
\newdimen\!ypos                 
\newdimen\!yprime               
\newdimen\!yshade               
\newdimen\!yshift               
\newdimen\!ytemp                
\newdimen\!ytop                 
\newdimen\!yunit                
\newdimen\!yyE                  
\newdimen\!yyM                  
\newdimen\!yyS                  
\newdimen\!yyloc                

\newif\if!axisvisible           
\newif\if!gridlinestoo          
\newif\if!keepPO                
\newif\if!placeaxislabel        
\newif\if!switch                
\newif\if!xswitch               

\newtoks\!axisLaBeL             
\newtoks\!keywordtoks           

\newwrite\!replotfile           

\newhelp\!keywordhelp{The keyword mentioned in the error message in unknown. 
Replace NEW KEYWORD in the indicated response by the keyword that 
should have been specified.}    

\!wlet\!!origin=\!xM                   
\!wlet\!!unit=\!uplength               
\!wlet\!Lresiduallength=\!dimenG       
\!wlet\!Rresiduallength=\!dimenF       
\!wlet\!axisLength=\!distacross        
\!wlet\!axisend=\!ydiff                
\!wlet\!axisstart=\!xdiff              
\!wlet\!axisxlevel=\!arclength         
\!wlet\!axisylevel=\!downlength        
\!wlet\!beta=\!dimenE                  
\!wlet\!gamma=\!dimenF                 
\!wlet\!shadexorigin=\!plotxorigin     
\!wlet\!shadeyorigin=\!plotyorigin     
\!wlet\!ticklength=\!xS                
\!wlet\!ticklocation=\!xE              
\!wlet\!ticklocationincr=\!yE          
\!wlet\!tickwidth=\!yS                 
\!wlet\!totalleaderlength=\!dimenE     
\!wlet\!xone=\!xprime                  
\!wlet\!xtwo=\!dxprime                 
\!wlet\!ySsave=\!yM                    
\!wlet\!ybB=\!yB                       
\!wlet\!ybC=\!yC                       
\!wlet\!ybE=\!yE                       
\!wlet\!ybM=\!yM                       
\!wlet\!ybS=\!yS                       
\!wlet\!ybpos=\!yyloc                  
\!wlet\!yone=\!yprime                  
\!wlet\!ytB=\!xB                       
\!wlet\!ytC=\!xC                       
\!wlet\!ytE=\!downlength               
\!wlet\!ytM=\!arclength                
\!wlet\!ytS=\!distacross               
\!wlet\!ytpos=\!xxloc                  
\!wlet\!ytwo=\!dyprime                 

\def\!zpt{0pt}                              
\!xunit=1pt
\!yunit=1pt
\!arearloc=\!xunit
\!areatloc=\!yunit
\!dshade=5pt
\!leaderlength=24in
\def\!tfs{256}                              
\def\!wmax{5.3pt}                           
\def\!wmin{2.7pt}                           
\!xaxislength=\!xunit
\!xpivot=\!zpt
\!yaxislength=\!yunit 
\!ypivot=\!zpt
  \!dimenA=50pt \def\!fiftypt{\the\!dimenA}     

\def\!rootten{3.162278pt}                   
\def\!tenAa{8.690286pt}                     
\def\!tenAc{2.773839pt}                     
\def\!tenAe{2.543275pt}                     

\def\!cosrotationangle{1}      
\def\!sinrotationangle{0}      
\def\!xpivotcoord{0}           
\def\!xref{0}                  
\def\!xshadesave{0}            
\def\!ypivotcoord{0}           
\def\!yref{0}                  
\def\!yshadesave{0}            
\def\!zero{0}                  

\let\wlog=\!!!wlog
\def\normalgraphs{%
  \longticklength=.4\baselineskip
  \shortticklength=.25\baselineskip
  \tickstovaluesleading=.25\baselineskip
  \valuestolabelleading=.8\baselineskip
  \linethickness=.4pt
  \stackleading=.17\baselineskip
  \headingtoplotskip=1.5\baselineskip
  \visibleaxes
  \ticksout
  \nogridlines
  \unloggedticks}
\def\setplotarea x from #1 to #2, y from #3 to #4 {%
  \!arealloc=\!M{#1}\!xunit \advance \!arealloc -\!xorigin
  \!areabloc=\!M{#3}\!yunit \advance \!areabloc -\!yorigin
  \!arearloc=\!M{#2}\!xunit \advance \!arearloc -\!xorigin
  \!areatloc=\!M{#4}\!yunit \advance \!areatloc -\!yorigin
  \!initinboundscheck
  \!xaxislength=\!arearloc  \advance\!xaxislength -\!arealloc
  \!yaxislength=\!areatloc  \advance\!yaxislength -\!areabloc
  \!plotheadingoffset=\!zpt
  \!dimenput {{\setbox0=\hbox{}\wd0=\!xaxislength\ht0=\!yaxislength\box0}}
     [bl] (\!arealloc,\!areabloc)}
\def\visibleaxes{%
  \def\!axisvisibility{\!axisvisibletrue}}

\def\!fixkeyword#1{%
  \errhelp=\!keywordhelp
  \errmessage{Unrecognized keyword `#1': \the\!keywordtoks{NEW KEYWORD}'}}

\!keywordtoks={enter `i\fixkeyword}

\def\fixkeyword#1{%
  \!nextkeyword#1 }

\def\axis {%
  \def\!nextkeyword##1 {%
    \expandafter\ifx\csname !axis##1\endcsname \relax
      \def\!next{\!fixkeyword{##1}}%
    \else
      \def\!next{\csname !axis##1\endcsname}%
    \fi
    \!next}%
  \!offset=\!zpt
  \!axisvisibility
  \!placeaxislabelfalse
  \!nextkeyword}

\def\!axisbottom{%
  \!axisylevel=\!areabloc
  \def\!tickxsign{0}%
  \def\!tickysign{-}%
  \def\!axissetup{\!axisxsetup}%
  \def\!axislabeltbrl{t}%
  \!nextkeyword}

\def\!axistop{%
  \!axisylevel=\!areatloc
  \def\!tickxsign{0}%
  \def\!tickysign{+}%
  \def\!axissetup{\!axisxsetup}%
  \def\!axislabeltbrl{b}%
  \!nextkeyword}

\def\!axisleft{%
  \!axisxlevel=\!arealloc
  \def\!tickxsign{-}%
  \def\!tickysign{0}%
  \def\!axissetup{\!axisysetup}%
  \def\!axislabeltbrl{r}%
  \!nextkeyword}

\def\!axisright{%
  \!axisxlevel=\!arearloc
  \def\!tickxsign{+}%
  \def\!tickysign{0}%
  \def\!axissetup{\!axisysetup}%
  \def\!axislabeltbrl{l}%
  \!nextkeyword}

\def\!axisshiftedto#1=#2 {%
  \if 0\!tickxsign
    \!axisylevel=\!M{#2}\!yunit
    \advance\!axisylevel -\!yorigin
  \else
    \!axisxlevel=\!M{#2}\!xunit
    \advance\!axisxlevel -\!xorigin
  \fi
  \!nextkeyword}

\def\!axisvisible{%
  \!axisvisibletrue  
  \!nextkeyword}

\def\!axisinvisible{%
  \!axisvisiblefalse
  \!nextkeyword}

\def\!axislabel#1 {%
  \!axisLaBeL={#1}%
  \!placeaxislabeltrue
  \!nextkeyword}

\expandafter\def\csname !axis/\endcsname{%
  \!axissetup 
  \if!placeaxislabel
    \!placeaxislabel
  \fi
  \if +\!tickysign 
    \!dimenA=\!axisylevel
    \advance\!dimenA \!offset 
    \advance\!dimenA -\!areatloc 
    \ifdim \!dimenA>\!plotheadingoffset
      \!plotheadingoffset=\!dimenA 
    \fi
  \fi}

\def\grid #1 #2 {%
  \!countA=#1\advance\!countA 1
  \axis bottom invisible ticks length <\!zpt> andacross quantity {\!countA} /
  \!countA=#2\advance\!countA 1
  \axis left   invisible ticks length <\!zpt> andacross quantity {\!countA} / }

\def\plotheading#1 {%
  \advance\!plotheadingoffset \headingtoplotskip
  \!dimenput {#1} [B] <.5\!xaxislength,\!plotheadingoffset>
    (\!arealloc,\!areatloc)}

\def\!axisxsetup{%
  \!axisxlevel=\!arealloc
  \!axisstart=\!arealloc
  \!axisend=\!arearloc
  \!axisLength=\!xaxislength
  \!!origin=\!xorigin
  \!!unit=\!xunit
  \!xswitchtrue
  \if!axisvisible 
    \!makeaxis
  \fi}

\def\!axisysetup{%
  \!axisylevel=\!areabloc
  \!axisstart=\!areabloc
  \!axisend=\!areatloc
  \!axisLength=\!yaxislength
  \!!origin=\!yorigin
  \!!unit=\!yunit
  \!xswitchfalse
  \if!axisvisible
    \!makeaxis
  \fi}

\def\!makeaxis{%
  \setbox\!boxA=\hbox{
    \beginpicture
      \!setdimenmode
      \setcoordinatesystem point at {\!zpt} {\!zpt}   
      \putrule from {\!zpt} {\!zpt} to
        {\!tickysign\!tickysign\!axisLength} 
        {\!tickxsign\!tickxsign\!axisLength}
    \endpicturesave <\!Xsave,\!Ysave>}%
    \wd\!boxA=\!zpt
    \!placetick\!axisstart}

\def\!placeaxislabel{%
  \advance\!offset \valuestolabelleading
  \if!xswitch
    \!dimenput {\the\!axisLaBeL} [\!axislabeltbrl]
      <.5\!axisLength,\!tickysign\!offset> (\!axisxlevel,\!axisylevel)
    \advance\!offset \!dp  
    \advance\!offset \!ht  
  \else
    \!dimenput {\the\!axisLaBeL} [\!axislabeltbrl]
      <\!tickxsign\!offset,.5\!axisLength> (\!axisxlevel,\!axisylevel)
  \fi
  \!axisLaBeL={}}

\def\arrow <#1> [#2,#3]{%
  \!ifnextchar<{\!arrow{#1}{#2}{#3}}{\!arrow{#1}{#2}{#3}<\!zpt,\!zpt> }}

\def\!arrow#1#2#3<#4,#5> from #6 #7 to #8 #9 {%
  \!xloc=\!M{#8}\!xunit   
  \!yloc=\!M{#9}\!yunit
  \!dxpos=\!xloc  \!dimenA=\!M{#6}\!xunit  \advance \!dxpos -\!dimenA
  \!dypos=\!yloc  \!dimenA=\!M{#7}\!yunit  \advance \!dypos -\!dimenA
  \let\!MAH=\!M
  \!setdimenmode
  \!xshift=#4\relax  \!yshift=#5\relax
  \!reverserotateonly\!xshift\!yshift
  \advance\!xshift\!xloc  \advance\!yshift\!yloc
  \!xS=-\!dxpos  \advance\!xS\!xshift
  \!yS=-\!dypos  \advance\!yS\!yshift
  \!start (\!xS,\!yS)
  \!ljoin (\!xshift,\!yshift)
  \!Pythag\!dxpos\!dypos\!arclength
  \!divide\!dxpos\!arclength\!dxpos  
  \!dxpos=32\!dxpos  \!removept\!dxpos\!!cos
  \!divide\!dypos\!arclength\!dypos  
  \!dypos=32\!dypos  \!removept\!dypos\!!sin
  \!halfhead{#1}{#2}{#3}
  \!halfhead{#1}{-#2}{-#3}
  \let\!M=\!MAH
  \ignorespaces}
  \def\!halfhead#1#2#3{%
    \!dimenC=-#1%
    \divide \!dimenC 2 
    \!dimenD=#2\!dimenC
    \!rotate(\!dimenC,\!dimenD)by(\!!cos,\!!sin)to(\!xM,\!yM)
    \!dimenC=-#1
    \!dimenD=#3\!dimenC
    \!dimenD=.5\!dimenD
    \!rotate(\!dimenC,\!dimenD)by(\!!cos,\!!sin)to(\!xE,\!yE)
    \!start (\!xshift,\!yshift)
    \advance\!xM\!xshift  \advance\!yM\!yshift
    \advance\!xE\!xshift  \advance\!yE\!yshift
    \!qjoin (\!xM,\!yM) (\!xE,\!yE) 
    \ignorespaces}

\def\betweenarrows #1#2 from #3 #4 to #5 #6 {%
  \!xloc=\!M{#3}\!xunit  \!xxloc=\!M{#5}\!xunit%
  \!yloc=\!M{#4}\!yunit  \!yyloc=\!M{#6}\!yunit%
  \!dxpos=\!xxloc  \advance\!dxpos by -\!xloc
  \!dypos=\!yyloc  \advance\!dypos by -\!yloc
  \advance\!xloc .5\!dxpos
  \advance\!yloc .5\!dypos
  \let\!MBA=\!M
  \!setdimenmode
  \ifdim\!dypos=\!zpt
    \ifdim\!dxpos<\!zpt \!dxpos=-\!dxpos \fi
    \put {\!lrarrows{\!dxpos}{#1}}#2{} at {\!xloc} {\!yloc}
  \else
    \ifdim\!dxpos=\!zpt
      \ifdim\!dypos<\!zpt \!dypos=-\!zpt \fi
      \put {\!udarrows{\!dypos}{#1}}#2{} at {\!xloc} {\!yloc}
    \fi
  \fi
  \let\!M=\!MBA
  \ignorespaces}

\def\!lrarrows#1#2{
  {\setbox\!boxA=\hbox{$\mkern-2mu\mathord-\mkern-2mu$}%
   \setbox\!boxB=\hbox{$\leftarrow$}\!dimenE=\ht\!boxB
   \setbox\!boxB=\hbox{}\ht\!boxB=2\!dimenE
   \hbox to #1{$\mathord\leftarrow\mkern-6mu
     \cleaders\copy\!boxA\hfil
     \mkern-6mu\mathord-$%
     \kern.4em $\vcenter{\box\!boxB}$$\vcenter{\hbox{#2}}$\kern.4em
     $\mathord-\mkern-6mu
     \cleaders\copy\!boxA\hfil
     \mkern-6mu\mathord\rightarrow$}}}

\def\!udarrows#1#2{
  {\setbox\!boxB=\hbox{#2}%
   \setbox\!boxA=\hbox to \wd\!boxB{\hss$\vert$\hss}%
   \!dimenE=\ht\!boxA \advance\!dimenE \dp\!boxA \divide\!dimenE 2
   \vbox to #1{\offinterlineskip
      \vskip .05556\!dimenE
      \hbox to \wd\!boxB{\hss$\mkern.4mu\uparrow$\hss}\vskip-\!dimenE
      \cleaders\copy\!boxA\vfil
      \vskip-\!dimenE\copy\!boxA
      \vskip\!dimenE\copy\!boxB\vskip.4em
      \copy\!boxA\vskip-\!dimenE
      \cleaders\copy\!boxA\vfil
      \vskip-\!dimenE \hbox to \wd\!boxB{\hss$\mkern.4mu\downarrow$\hss}
      \vskip .05556\!dimenE}}}

\def\putbar#1breadth <#2> from #3 #4 to #5 #6 {%
  \!xloc=\!M{#3}\!xunit  \!xxloc=\!M{#5}\!xunit%
  \!yloc=\!M{#4}\!yunit  \!yyloc=\!M{#6}\!yunit%
  \!dypos=\!yyloc  \advance\!dypos by -\!yloc
  \!dimenI=#2  
  \ifdim \!dimenI=\!zpt 
    \putrule#1from {#3} {#4} to {#5} {#6} 
  \else 
    \let\!MBar=\!M
    \!setdimenmode 
    \divide\!dimenI 2
    \ifdim \!dypos=\!zpt             
      \advance \!yloc -\!dimenI 
      \advance \!yyloc \!dimenI
    \else
      \advance \!xloc -\!dimenI 
      \advance \!xxloc \!dimenI
    \fi
    \putrectangle#1corners at {\!xloc} {\!yloc} and {\!xxloc} {\!yyloc}
    \let\!M=\!MBar 
  \fi
  \ignorespaces}

\def\setbars#1breadth <#2> baseline at #3 = #4 {%
  \edef\!barshift{#1}%
  \edef\!barbreadth{#2}%
  \edef\!barorientation{#3}%
  \edef\!barbaseline{#4}%
  \def\!bardobaselabel{\!bardoendlabel}%
  \def\!bardoendlabel{\!barfinish}%
  \let\!drawcurve=\!barcurve
  \!setbars}
\def\!setbars{%
  \futurelet\!nextchar\!!setbars}
\def\!!setbars{%
  \if b\!nextchar
    \def\!!!setbars{\!setbarsbget}%
  \else 
    \if e\!nextchar
      \def\!!!setbars{\!setbarseget}%
    \else
      \def\!!!setbars{\relax}%
    \fi
  \fi
  \!!!setbars}
\def\!setbarsbget baselabels (#1) {%
  \def\!barbaselabelorientation{#1}%
  \def\!bardobaselabel{\!!bardobaselabel}%
  \!setbars}
\def\!setbarseget endlabels (#1) {%
  \edef\!barendlabelorientation{#1}%
  \def\!bardoendlabel{\!!bardoendlabel}%
  \!setbars}

\def\!barcurve #1 #2 {%
  \if y\!barorientation
    \def\!basexarg{#1}%
    \def\!baseyarg{\!barbaseline}%
  \else
    \def\!basexarg{\!barbaseline}%
    \def\!baseyarg{#2}%
  \fi
  \expandafter\putbar\!barshift breadth <\!barbreadth> from {\!basexarg}
    {\!baseyarg} to {#1} {#2}
  \def\!endxarg{#1}%
  \def\!endyarg{#2}%
  \!bardobaselabel}

\def\!!bardobaselabel "#1" {%
  \put {#1}\!barbaselabelorientation{} at {\!basexarg} {\!baseyarg}
  \!bardoendlabel}
 
\def\!!bardoendlabel "#1" {%
  \put {#1}\!barendlabelorientation{} at {\!endxarg} {\!endyarg}
  \!barfinish}

\def\!barfinish{%
  \!ifnextchar/{\!finish}{\!barcurve}}

\def\putrectangle{%
  \!ifnextchar<{\!putrectangle}{\!putrectangle<\!zpt,\!zpt> }}
\def\!putrectangle<#1,#2> corners at #3 #4 and #5 #6 {%
  \!xone=\!M{#3}\!xunit  \!xtwo=\!M{#5}\!xunit%
  \!yone=\!M{#4}\!yunit  \!ytwo=\!M{#6}\!yunit%
  \ifdim \!xtwo<\!xone
    \!dimenI=\!xone  \!xone=\!xtwo  \!xtwo=\!dimenI
  \fi
  \ifdim \!ytwo<\!yone
    \!dimenI=\!yone  \!yone=\!ytwo  \!ytwo=\!dimenI
  \fi
  \!dimenI=#1\relax  \advance\!xone\!dimenI  \advance\!xtwo\!dimenI
  \!dimenI=#2\relax  \advance\!yone\!dimenI  \advance\!ytwo\!dimenI
  \let\!MRect=\!M
  \!setdimenmode
  \!shaderectangle
  \!dimenI=.5\linethickness
  \advance \!xone  -\!dimenI
  \advance \!xtwo   \!dimenI
  \putrule from {\!xone} {\!yone} to {\!xtwo} {\!yone} 
  \putrule from {\!xone} {\!ytwo} to {\!xtwo} {\!ytwo} 
  \advance \!xone   \!dimenI
  \advance \!xtwo  -\!dimenI%
  \advance \!yone  -\!dimenI
  \advance \!ytwo   \!dimenI
  \putrule from {\!xone} {\!yone} to {\!xone} {\!ytwo} 
  \putrule from {\!xtwo} {\!yone} to {\!xtwo} {\!ytwo} 
  \let\!M=\!MRect
  \ignorespaces}
 
\def\shaderectanglesoff{%
  \def\!shaderectangle{}%
  \ignorespaces}

\shaderectanglesoff
 
\def\!!shaderectangle{%
  \!dimenA=\!xtwo  \advance \!dimenA -\!xone
  \!dimenB=\!ytwo  \advance \!dimenB -\!yone
  \ifdim \!dimenA<\!dimenB
    \!startvshade (\!xone,\!yone,\!ytwo)
    \!lshade      (\!xtwo,\!yone,\!ytwo)
  \else
    \!starthshade (\!yone,\!xone,\!xtwo)
    \!lshade      (\!ytwo,\!xone,\!xtwo)
  \fi
  \ignorespaces}
  
\def\frame{%
  \!ifnextchar<{\!frame}{\!frame<\!zpt> }}
\long\def\!frame<#1> #2{%
  \beginpicture
    \setcoordinatesystem units <1pt,1pt> point at 0 0 
    \put {#2} [Bl] at 0 0 
    \!dimenA=#1\relax
    \!dimenB=\!wd \advance \!dimenB \!dimenA
    \!dimenC=\!ht \advance \!dimenC \!dimenA
    \!dimenD=\!dp \advance \!dimenD \!dimenA
    \let\!MFr=\!M
    \!setdimenmode
    \putrectangle corners at {-\!dimenA} {-\!dimenD} and {\!dimenB} {\!dimenC}
    \!setcoordmode
    \let\!M=\!MFr
  \endpicture
  \ignorespaces}
 
\def\rectangle <#1> <#2> {%
  \setbox0=\hbox{}\wd0=#1\ht0=#2\frame {\box0}}

\def\plot{%
  \!ifnextchar"{\!plotfromfile}{\!drawcurve}}
\def\!plotfromfile"#1"{%
  \expandafter\!drawcurve \input #1 /}

\def\setquadratic{%
  \let\!drawcurve=\!qcurve
  \let\!!Shade=\!!qShade
  \let\!!!Shade=\!!!qShade}

\def\setlinear{%
  \let\!drawcurve=\!lcurve
  \let\!!Shade=\!!lShade
  \let\!!!Shade=\!!!lShade}

\def\sethistograms{%
  \let\!drawcurve=\!hcurve}

\def\!qcurve #1 #2 {%
  \!start (#1,#2)
  \!Qjoin}
\def\!Qjoin#1 #2 #3 #4 {%
  \!qjoin (#1,#2) (#3,#4)             
  \!ifnextchar/{\!finish}{\!Qjoin}}

\def\!lcurve #1 #2 {%
  \!start (#1,#2)
  \!Ljoin}
\def\!Ljoin#1 #2 {%
  \!ljoin (#1,#2)                    
  \!ifnextchar/{\!finish}{\!Ljoin}}

\def\!finish/{\ignorespaces}

\def\!hcurve #1 #2 {%
  \edef\!hxS{#1}%
  \edef\!hyS{#2}%
  \!hjoin}
\def\!hjoin#1 #2 {%
  \putrectangle corners at {\!hxS} {\!hyS} and {#1} {#2}
  \edef\!hxS{#1}%
  \!ifnextchar/{\!finish}{\!hjoin}}

\def\vshade #1 #2 #3 {%
  \!startvshade (#1,#2,#3)
  \!Shadewhat}

\def\hshade #1 #2 #3 {%
  \!starthshade (#1,#2,#3)
  \!Shadewhat}

\def\!Shadewhat{%
  \futurelet\!nextchar\!Shade}
\def\!Shade{%
  \if <\!nextchar
    \def\!nextShade{\!!Shade}%
  \else
    \if /\!nextchar
      \def\!nextShade{\!finish}%
    \else
      \def\!nextShade{\!!!Shade}%
    \fi
  \fi
  \!nextShade}
\def\!!lShade<#1> #2 #3 #4 {%
  \!lshade <#1> (#2,#3,#4)                 
  \!Shadewhat}
\def\!!!lShade#1 #2 #3 {%
  \!lshade (#1,#2,#3)
  \!Shadewhat} 
\def\!!qShade<#1> #2 #3 #4 #5 #6 #7 {%
  \!qshade <#1> (#2,#3,#4) (#5,#6,#7)      
  \!Shadewhat}
\def\!!!qShade#1 #2 #3 #4 #5 #6 {%
  \!qshade (#1,#2,#3) (#4,#5,#6)
  \!Shadewhat} 

\setlinear

\def\setdashpattern <#1>{%
  \def\!Flist{}\def\!Blist{}\def\!UDlist{}%
  \!countA=0
  \!ecfor\!item:=#1\do{%
    \!dimenA=\!item\relax
    \expandafter\!rightappend\the\!dimenA\withCS{\\}\to\!UDlist%
    \advance\!countA  1
    \ifodd\!countA
      \expandafter\!rightappend\the\!dimenA\withCS{\!Rule}\to\!Flist%
      \expandafter\!leftappend\the\!dimenA\withCS{\!Rule}\to\!Blist%
    \else 
      \expandafter\!rightappend\the\!dimenA\withCS{\!Skip}\to\!Flist%
      \expandafter\!leftappend\the\!dimenA\withCS{\!Skip}\to\!Blist%
    \fi}%
  \!leaderlength=\!zpt
  \def\!Rule##1{\advance\!leaderlength  ##1}%
  \def\!Skip##1{\advance\!leaderlength  ##1}%
  \!Flist%
  \ifdim\!leaderlength>\!zpt 
  \else
    \def\!Flist{\!Skip{24in}}\def\!Blist{\!Skip{24in}}\ignorespaces
    \def\!UDlist{\\{\!zpt}\\{24in}}\ignorespaces
    \!leaderlength=24in
  \fi
  \!dashingon}

\def\!dashingon{%
  \def\!advancedashing{\!!advancedashing}%
  \def\!drawlinearsegment{\!lineardashed}%
  \def\!puthline{\!putdashedhline}%
  \def\!putvline{\!putdashedvline}%
  \ignorespaces}%
\def\!dashingoff{%
  \def\!advancedashing{\relax}%
  \def\!drawlinearsegment{\!linearsolid}%
  \def\!puthline{\!putsolidhline}%
  \def\!putvline{\!putsolidvline}%
  \ignorespaces}

\def\setdots{%
  \!ifnextchar<{\!setdots}{\!setdots<5pt>}}
\def\!setdots<#1>{%
  \!dimenB=#1\advance\!dimenB -\plotsymbolspacing
  \ifdim\!dimenB<\!zpt
    \!dimenB=\!zpt
  \fi
\setdashpattern <\plotsymbolspacing,\!dimenB>}
 
\def\setdotsnear <#1> for <#2>{%
  \!dimenB=#2\relax  \advance\!dimenB -.05pt  
  \!dimenC=#1\relax  \!countA=\!dimenC 
  \!dimenD=\!dimenB  \advance\!dimenD .5\!dimenC  \!countB=\!dimenD
  \divide \!countB  \!countA
  \ifnum 1>\!countB 
    \!countB=1
  \fi
  \divide\!dimenB  \!countB
  \setdots <\!dimenB>}
 
\def\setdashes{%
  \!ifnextchar<{\!setdashes}{\!setdashes<5pt>}}
\def\!setdashes<#1>{\setdashpattern <#1,#1>}
 
\def\setdashesnear <#1> for <#2>{%
  \!dimenB=#2\relax  
  \!dimenC=#1\relax  \!countA=\!dimenC 
  \!dimenD=\!dimenB  \advance\!dimenD .5\!dimenC  \!countB=\!dimenD
  \divide \!countB  \!countA
  \ifodd \!countB 
  \else 
    \advance \!countB  1
  \fi
  \divide\!dimenB  \!countB
  \setdashes <\!dimenB>}
 
\def\setsolid{%
  \def\!Flist{\!Rule{24in}}\def\!Blist{\!Rule{24in}}%
  \def\!UDlist{\\{24in}\\{\!zpt}}%
  \!dashingoff}  
\setsolid

\def\!divide#1#2#3{%
  \!dimenB=#1
  \!dimenC=#2
  \!dimenD=\!dimenB
  \divide \!dimenD \!dimenC
  \!dimenA=\!dimenD
  \multiply\!dimenD \!dimenC
  \advance\!dimenB -\!dimenD
  \!dimenD=\!dimenC
    \ifdim\!dimenD<\!zpt \!dimenD=-\!dimenD 
  \fi
  \ifdim\!dimenD<64pt
    \!divstep[\!tfs]\!divstep[\!tfs]%
  \else 
    \!!divide
  \fi
  #3=\!dimenA\ignorespaces}

\def\!!divide{%
  \ifdim\!dimenD<256pt
    \!divstep[64]\!divstep[32]\!divstep[32]%
  \else 
    \!divstep[8]\!divstep[8]\!divstep[8]\!divstep[8]\!divstep[8]%
    \!dimenA=2\!dimenA
  \fi}

\def\!divstep[#1]{
  \!dimenB=#1\!dimenB
  \!dimenD=\!dimenB
    \divide \!dimenD by \!dimenC
  \!dimenA=#1\!dimenA
    \advance\!dimenA by \!dimenD%
  \multiply\!dimenD by \!dimenC
    \advance\!dimenB by -\!dimenD}
 
\def\Divide <#1> by <#2> forming <#3> {%
  \!divide{#1}{#2}{#3}}

\def\ellipticalarc axes ratio #1:#2 #3 degrees from #4 #5 center at #6 #7 {%
  \!angle=#3pt\relax
  \ifdim\!angle>\!zpt 
    \def\!sign{}
  \else 
    \def\!sign{-}\!angle=-\!angle
  \fi
  \!xxloc=\!M{#6}\!xunit
  \!yyloc=\!M{#7}\!yunit     
  \!xxS=\!M{#4}\!xunit
  \!yyS=\!M{#5}\!yunit
  \advance\!xxS -\!xxloc
  \advance\!yyS -\!yyloc
  \!divide\!xxS{#1pt}\!xxS 
  \!divide\!yyS{#2pt}\!yyS 
  \let\!MC=\!M
  \!setdimenmode
  \!xS=#1\!xxS  \advance\!xS\!xxloc
  \!yS=#2\!yyS  \advance\!yS\!yyloc
  \!start (\!xS,\!yS)%
  \!loop\ifdim\!angle>14.9999pt
    \!rotate(\!xxS,\!yyS)by(\!cos,\!sign\!sin)to(\!xxM,\!yyM) 
    \!rotate(\!xxM,\!yyM)by(\!cos,\!sign\!sin)to(\!xxE,\!yyE)
    \!xM=#1\!xxM  \advance\!xM\!xxloc  \!yM=#2\!yyM  \advance\!yM\!yyloc
    \!xE=#1\!xxE  \advance\!xE\!xxloc  \!yE=#2\!yyE  \advance\!yE\!yyloc
    \!qjoin (\!xM,\!yM) (\!xE,\!yE)
    \!xxS=\!xxE  \!yyS=\!yyE 
    \advance \!angle -15pt
  \repeat
  \ifdim\!angle>\!zpt
    \!angle=100.53096\!angle
    \divide \!angle 360 
    \!sinandcos\!angle\!!sin\!!cos
    \!rotate(\!xxS,\!yyS)by(\!!cos,\!sign\!!sin)to(\!xxM,\!yyM) 
    \!rotate(\!xxM,\!yyM)by(\!!cos,\!sign\!!sin)to(\!xxE,\!yyE)
    \!xM=#1\!xxM  \advance\!xM\!xxloc  \!yM=#2\!yyM  \advance\!yM\!yyloc
    \!xE=#1\!xxE  \advance\!xE\!xxloc  \!yE=#2\!yyE  \advance\!yE\!yyloc
    \!qjoin (\!xM,\!yM) (\!xE,\!yE)
  \fi
  \let\!M=\!MC
  \ignorespaces}

\def\!rotate(#1,#2)by(#3,#4)to(#5,#6){%
  \!dimenA=#3#1\advance \!dimenA -#4#2
  \!dimenB=#3#2\advance \!dimenB  #4#1
  \divide \!dimenA 32  \divide \!dimenB 32 
  #5=\!dimenA  #6=\!dimenB
  \ignorespaces}
\def\!sin{4.17684}
\def\!cos{31.72624}

\def\!sinandcos#1#2#3{%
 \!dimenD=#1
 \!dimenA=\!dimenD
 \!dimenB=32pt
 \!removept\!dimenD\!value
 \!dimenC=\!dimenD
 \!dimenC=\!value\!dimenC \divide\!dimenC by 64 
 \advance\!dimenB by -\!dimenC
 \!dimenC=\!value\!dimenC \divide\!dimenC by 96 
 \advance\!dimenA by -\!dimenC
 \!dimenC=\!value\!dimenC \divide\!dimenC by 128 
 \advance\!dimenB by \!dimenC%
 \!removept\!dimenA#2
 \!removept\!dimenB#3
 \ignorespaces}

\def\putrule#1from #2 #3 to #4 #5 {%
  \!xloc=\!M{#2}\!xunit  \!xxloc=\!M{#4}\!xunit%
  \!yloc=\!M{#3}\!yunit  \!yyloc=\!M{#5}\!yunit%
  \!dxpos=\!xxloc  \advance\!dxpos by -\!xloc
  \!dypos=\!yyloc  \advance\!dypos by -\!yloc
  \ifdim\!dypos=\!zpt
    \def\!!Line{\!puthline{#1}}\ignorespaces
  \else
    \ifdim\!dxpos=\!zpt
      \def\!!Line{\!putvline{#1}}\ignorespaces
    \else 
       \def\!!Line{}
    \fi
  \fi
  \let\!ML=\!M
  \!setdimenmode
  \!!Line%
  \let\!M=\!ML
  \ignorespaces}

\def\!putsolidhline#1{%
  \ifdim\!dxpos>\!zpt 
    \put{\!hline\!dxpos}#1[l] at {\!xloc} {\!yloc}
  \else 
    \put{\!hline{-\!dxpos}}#1[l] at {\!xxloc} {\!yyloc}
  \fi
  \ignorespaces}
 
\def\!putsolidvline#1{%
  \ifdim\!dypos>\!zpt 
    \put{\!vline\!dypos}#1[b] at {\!xloc} {\!yloc}
  \else 
    \put{\!vline{-\!dypos}}#1[b] at {\!xxloc} {\!yyloc}
  \fi
  \ignorespaces}
 
\def\!hline#1{\hbox to #1{\leaders \hrule height\linethickness\hfill}}
\def\!vline#1{\vbox to #1{\leaders \vrule width\linethickness\vfill}}

\def\!putdashedhline#1{%
  \ifdim\!dxpos>\!zpt 
    \!DLsetup\!Flist\!dxpos
    \put{\hbox to \!totalleaderlength{\!hleaders}\!hpartialpattern\!Rtrunc}
      #1[l] at {\!xloc} {\!yloc} 
  \else 
    \!DLsetup\!Blist{-\!dxpos}
    \put{\!hpartialpattern\!Ltrunc\hbox to \!totalleaderlength{\!hleaders}}
      #1[r] at {\!xloc} {\!yloc} 
  \fi
  \ignorespaces}
 
\def\!putdashedvline#1{%
  \!dypos=-\!dypos
  \ifdim\!dypos>\!zpt 
    \!DLsetup\!Flist\!dypos 
    \put{\vbox{\vbox to \!totalleaderlength{\!vleaders}
      \!vpartialpattern\!Rtrunc}}#1[t] at {\!xloc} {\!yloc} 
  \else 
    \!DLsetup\!Blist{-\!dypos}
    \put{\vbox{\!vpartialpattern\!Ltrunc
      \vbox to \!totalleaderlength{\!vleaders}}}#1[b] at {\!xloc} {\!yloc} 
  \fi
  \ignorespaces}

\def\!DLsetup#1#2{
  \let\!RSlist=#1
  \!countB=#2
  \!countA=\!leaderlength
  \divide\!countB by \!countA
  \!totalleaderlength=\!countB\!leaderlength
  \!Rresiduallength=#2%
  \advance \!Rresiduallength by -\!totalleaderlength
  \!Lresiduallength=\!leaderlength
  \advance \!Lresiduallength by -\!Rresiduallength
  \ignorespaces}
 
\def\!hleaders{%
  \def\!Rule##1{\vrule height\linethickness width##1}%
  \def\!Skip##1{\hskip##1}%
  \leaders\hbox{\!RSlist}\hfill}
 
\def\!hpartialpattern#1{%
  \!dimenA=\!zpt \!dimenB=\!zpt 
  \def\!Rule##1{#1{##1}\vrule height\linethickness width\!dimenD}%
  \def\!Skip##1{#1{##1}\hskip\!dimenD}%
  \!RSlist}
 
\def\!vleaders{%
  \def\!Rule##1{\hrule width\linethickness height##1}%
  \def\!Skip##1{\vskip##1}%
  \leaders\vbox{\!RSlist}\vfill}
 
\def\!vpartialpattern#1{%
  \!dimenA=\!zpt \!dimenB=\!zpt 
  \def\!Rule##1{#1{##1}\hrule width\linethickness height\!dimenD}%
  \def\!Skip##1{#1{##1}\vskip\!dimenD}%
  \!RSlist}
 
\def\!Rtrunc#1{\!trunc{#1}>\!Rresiduallength}
\def\!Ltrunc#1{\!trunc{#1}<\!Lresiduallength}
 
\def\!trunc#1#2#3{%
  \!dimenA=\!dimenB         
  \advance\!dimenB by #1%
  \!dimenD=\!dimenB  \ifdim\!dimenD#2#3\!dimenD=#3\fi
  \!dimenC=\!dimenA  \ifdim\!dimenC#2#3\!dimenC=#3\fi
  \advance \!dimenD by -\!dimenC}

\def\!start (#1,#2){%
  \!plotxorigin=\!xorigin  \advance \!plotxorigin by \!plotsymbolxshift
  \!plotyorigin=\!yorigin  \advance \!plotyorigin by \!plotsymbolyshift
  \!xS=\!M{#1}\!xunit \!yS=\!M{#2}\!yunit
  \!rotateaboutpivot\!xS\!yS
  \!copylist\!UDlist\to\!!UDlist
  \!getnextvalueof\!downlength\from\!!UDlist
  \!distacross=\!zpt
  \!intervalno=0 
  \global\totalarclength=\!zpt
  \ignorespaces}

\def\!ljoin (#1,#2){%
  \advance\!intervalno by 1
  \!xE=\!M{#1}\!xunit \!yE=\!M{#2}\!yunit
  \!rotateaboutpivot\!xE\!yE
  \!xdiff=\!xE \advance \!xdiff by -\!xS
  \!ydiff=\!yE \advance \!ydiff by -\!yS
  \!Pythag\!xdiff\!ydiff\!arclength
  \global\advance \totalarclength by \!arclength%
  \!drawlinearsegment
  \!xS=\!xE \!yS=\!yE
  \ignorespaces}

\def\!linearsolid{%
  \!npoints=\!arclength
  \!countA=\plotsymbolspacing
  \divide\!npoints by \!countA
  \ifnum \!npoints<1 
    \!npoints=1 
  \fi
  \divide\!xdiff by \!npoints
  \divide\!ydiff by \!npoints
  \!xpos=\!xS \!ypos=\!yS
  \loop\ifnum\!npoints>-1
    \!plotifinbounds
    \advance \!xpos by \!xdiff
    \advance \!ypos by \!ydiff
    \advance \!npoints by -1
  \repeat
  \ignorespaces}

\def\!lineardashed{%
  \ifdim\!distacross>\!arclength
    \advance \!distacross by -\!arclength  
  \else
    \loop\ifdim\!distacross<\!arclength
      \!divide\!distacross\!arclength\!dimenA
      \!removept\!dimenA\!t
      \!xpos=\!t\!xdiff \advance \!xpos by \!xS
      \!ypos=\!t\!ydiff \advance \!ypos by \!yS
      \!plotifinbounds
      \advance\!distacross by \plotsymbolspacing
      \!advancedashing
    \repeat  
    \advance \!distacross by -\!arclength
  \fi
  \ignorespaces}

\def\!!advancedashing{%
  \advance\!downlength by -\plotsymbolspacing
  \ifdim \!downlength>\!zpt
  \else
    \advance\!distacross by \!downlength
    \!getnextvalueof\!uplength\from\!!UDlist
    \advance\!distacross by \!uplength
    \!getnextvalueof\!downlength\from\!!UDlist
  \fi}

\def\inboundscheckoff{%
  \def\!plotifinbounds{\!plot(\!xpos,\!ypos)}%
  \def\!initinboundscheck{\relax}\ignorespaces}
 
\inboundscheckoff
 
\def\!!plotifinbounds{%
  \ifdim \!xpos<\!checkleft
  \else
    \ifdim \!xpos>\!checkright
    \else
      \ifdim \!ypos<\!checkbot
      \else
         \ifdim \!ypos>\!checktop
         \else
           \!plot(\!xpos,\!ypos)
         \fi 
      \fi
    \fi
  \fi}

\def\!!initinboundscheck{%
  \!checkleft=\!arealloc     \advance\!checkleft by \!xorigin
  \!checkright=\!arearloc    \advance\!checkright by \!xorigin
  \!checkbot=\!areabloc      \advance\!checkbot by \!yorigin
  \!checktop=\!areatloc      \advance\!checktop by \!yorigin}

\def\!logten#1#2{%
  \expandafter\!!logten#1\!nil
  \!removept\!dimenF#2%
  \ignorespaces}

\def\!!logten#1#2\!nil{%
  \if -#1%
    \!dimenF=\!zpt
    \def\!next{\ignorespaces}%
  \else
    \if +#1%
      \def\!next{\!!logten#2\!nil}%
    \else
      \if .#1%
        \def\!next{\!!logten0.#2\!nil}%
      \else
        \def\!next{\!!!logten#1#2..\!nil}%
      \fi
    \fi
  \fi
  \!next}

\def\!!!logten#1#2.#3.#4\!nil{%
  \!dimenF=1pt 
  \if 0#1%
    \!!logshift#3pt 
  \else 
    \!logshift#2/
    \!dimenE=#1.#2#3pt 
  \fi 
  \ifdim \!dimenE<\!rootten
    \multiply \!dimenE 10 
    \advance  \!dimenF -1pt
  \fi
  \!dimenG=\!dimenE
    \advance\!dimenG 10pt
  \advance\!dimenE -10pt 
  \multiply\!dimenE 10 
  \!divide\!dimenE\!dimenG\!dimenE
  \!removept\!dimenE\!t
  \!dimenG=\!t\!dimenE
  \!removept\!dimenG\!tt
  \!dimenH=\!tt\!tenAe
    \divide\!dimenH 100
  \advance\!dimenH \!tenAc
  \!dimenH=\!tt\!dimenH
    \divide\!dimenH 100   
  \advance\!dimenH \!tenAa
  \!dimenH=\!t\!dimenH
    \divide\!dimenH 100 
  \advance\!dimenF \!dimenH}

\def\!logshift#1{%
  \if #1/%
    \def\!next{\ignorespaces}%
  \else
    \advance\!dimenF 1pt 
    \def\!next{\!logshift}%
  \fi 
  \!next}
 
 \def\!!logshift#1{%
   \advance\!dimenF -1pt
   \if 0#1%
     \def\!next{\!!logshift}%
   \else
     \if p#1%
       \!dimenF=1pt
       \def\!next{\!dimenE=1p}%
     \else
       \def\!next{\!dimenE=#1.}%
     \fi
   \fi
   \!next}

\def\beginpicture{%
  \setbox\!picbox=\hbox\bgroup%
  \!xleft=\maxdimen  
  \!xright=-\maxdimen
  \!ybot=\maxdimen
  \!ytop=-\maxdimen}
 
\def\endpicture{%
  \ifdim\!xleft=\maxdimen
    \!xleft=\!zpt \!xright=\!zpt \!ybot=\!zpt \!ytop=\!zpt 
  \fi
  \global\!Xleft=\!xleft \global\!Xright=\!xright
  \global\!Ybot=\!ybot \global\!Ytop=\!ytop
  \egroup%
  \ht\!picbox=\!Ytop  \dp\!picbox=-\!Ybot
  \ifdim\!Ybot>\!zpt
  \else 
    \ifdim\!Ytop<\!zpt
      \!Ybot=\!Ytop
    \else
      \!Ybot=\!zpt
    \fi
  \fi
  \hbox{\kern-\!Xleft\lower\!Ybot\box\!picbox\kern\!Xright}}
 
\def\endpicturesave <#1,#2>{%
  \endpicture \global #1=\!Xleft \global #2=\!Ybot \ignorespaces}

\def\setcoordinatesystem{%
  \!ifnextchar{u}{\!getlengths }
    {\!getlengths units <\!xunit,\!yunit>}}
\def\!getlengths units <#1,#2>{%
  \!xunit=#1\relax
  \!yunit=#2\relax
  \!ifcoordmode 
    \let\!SCnext=\!SCccheckforRP
  \else
    \let\!SCnext=\!SCdcheckforRP
  \fi
  \!SCnext}
\def\!SCccheckforRP{%
  \!ifnextchar{p}{\!cgetreference }
    {\!cgetreference point at {\!xref} {\!yref} }}
\def\!cgetreference point at #1 #2 {%
  \edef\!xref{#1}\edef\!yref{#2}%
  \!xorigin=\!xref\!xunit  \!yorigin=\!yref\!yunit  
  \!initinboundscheck 
  \ignorespaces}
\def\!SCdcheckforRP{%
  \!ifnextchar{p}{\!dgetreference}%
    {\ignorespaces}}
\def\!dgetreference point at #1 #2 {%
  \!xorigin=#1\relax  \!yorigin=#2\relax
  \ignorespaces}

\long\def\put#1#2 at #3 #4 {%
  \!setputobject{#1}{#2}%
  \!xpos=\!M{#3}\!xunit  \!ypos=\!M{#4}\!yunit  
  \!rotateaboutpivot\!xpos\!ypos%
  \advance\!xpos -\!xorigin  \advance\!xpos -\!xshift
  \advance\!ypos -\!yorigin  \advance\!ypos -\!yshift
  \kern\!xpos\raise\!ypos\box\!putobject\kern-\!xpos%
  \!doaccounting\ignorespaces}
 
\long\def\multiput #1#2 at {%
  \!setputobject{#1}{#2}%
  \!ifnextchar"{\!putfromfile}{\!multiput}}
\def\!putfromfile"#1"{%
  \expandafter\!multiput \input #1 /}
\def\!multiput{%
  \futurelet\!nextchar\!!multiput}
\def\!!multiput{%
  \if *\!nextchar
    \def\!nextput{\!alsoby}%
  \else
    \if /\!nextchar
      \def\!nextput{\!finishmultiput}%
    \else
      \def\!nextput{\!alsoat}%
    \fi
  \fi
  \!nextput}
\def\!finishmultiput/{%
  \setbox\!putobject=\hbox{}%
  \ignorespaces}
 
\def\!alsoat#1 #2 {%
  \!xpos=\!M{#1}\!xunit  \!ypos=\!M{#2}\!yunit  
  \!rotateaboutpivot\!xpos\!ypos%
  \advance\!xpos -\!xorigin  \advance\!xpos -\!xshift
  \advance\!ypos -\!yorigin  \advance\!ypos -\!yshift
  \kern\!xpos\raise\!ypos\copy\!putobject\kern-\!xpos%
  \!doaccounting
  \!multiput}
 
\def\!alsoby*#1 #2 #3 {%
  \!dxpos=\!M{#2}\!xunit \!dypos=\!M{#3}\!yunit 
  \!rotateonly\!dxpos\!dypos
  \!ntemp=#1%
  \!!loop\ifnum\!ntemp>0
    \advance\!xpos by \!dxpos  \advance\!ypos by \!dypos
    \kern\!xpos\raise\!ypos\copy\!putobject\kern-\!xpos%
    \advance\!ntemp by -1
  \repeat
  \!doaccounting 
  \!multiput}
 
\def\accountingon{\def\!doaccounting{\!!doaccounting}\ignorespaces}

\accountingon
\def\!!doaccounting{%
  \!xtemp=\!xpos  
  \!ytemp=\!ypos
  \ifdim\!xtemp<\!xleft 
     \!xleft=\!xtemp 
  \fi
  \advance\!xtemp by  \!wd 
  \ifdim\!xright<\!xtemp 
    \!xright=\!xtemp
  \fi
  \advance\!ytemp by -\!dp
  \ifdim\!ytemp<\!ybot  
    \!ybot=\!ytemp
  \fi
  \advance\!ytemp by  \!dp
  \advance\!ytemp by  \!ht 
  \ifdim\!ytemp>\!ytop  
    \!ytop=\!ytemp  
  \fi}
 
\long\def\!setputobject#1#2{%
  \setbox\!putobject=\hbox{#1}%
  \!ht=\ht\!putobject  \!dp=\dp\!putobject  \!wd=\wd\!putobject
  \wd\!putobject=\!zpt
  \!xshift=.5\!wd   \!yshift=.5\!ht   \advance\!yshift by -.5\!dp
  \edef\!putorientation{#2}%
  \expandafter\!SPOreadA\!putorientation[]\!nil%
  \expandafter\!SPOreadB\!putorientation<\!zpt,\!zpt>\!nil\ignorespaces}
 
\def\!SPOreadA#1[#2]#3\!nil{\!etfor\!orientation:=#2\do\!SPOreviseshift}
 
\def\!SPOreadB#1<#2,#3>#4\!nil{\advance\!xshift by -#2\advance\!yshift by -#3}
 
\def\!SPOreviseshift{%
  \if l\!orientation 
    \!xshift=\!zpt
  \else 
    \if r\!orientation 
      \!xshift=\!wd
    \else 
      \if b\!orientation
        \!yshift=-\!dp
      \else 
        \if B\!orientation 
          \!yshift=\!zpt
        \else 
          \if t\!orientation 
            \!yshift=\!ht
          \fi 
        \fi
      \fi
    \fi
  \fi}

\long\def\!dimenput#1#2(#3,#4){%
  \!setputobject{#1}{#2}%
  \!xpos=#3\advance\!xpos by -\!xshift
  \!ypos=#4\advance\!ypos by -\!yshift
  \kern\!xpos\raise\!ypos\box\!putobject\kern-\!xpos%
  \!doaccounting\ignorespaces}

\def\!setdimenmode{%
  \let\!M=\!M!!\ignorespaces}
\def\!setcoordmode{%
  \let\!M=\!M!\ignorespaces}
\def\!ifcoordmode{%
  \ifx \!M \!M!}
\def\!ifdimenmode{%
  \ifx \!M \!M!!}
\def\!M!#1#2{#1#2} 
\def\!M!!#1#2{#1}
\!setcoordmode
\let\setdimensionmode=\!setdimenmode
\let\setcoordinatemode=\!setcoordmode

\def\stack{%
  \!ifnextchar[{\!stack}{\!stack[c]}}
\def\!stack[#1]{%
  \let\!lglue=\hfill \let\!rglue=\hfill
  \expandafter\let\csname !#1glue\endcsname=\relax
  \!ifnextchar<{\!!stack}{\!!stack<\stackleading>}}
\def\!!stack<#1>#2{%
  \vbox{\def\!valueslist{}\!ecfor\!value:=#2\do{%
    \expandafter\!rightappend\!value\withCS{\\}\to\!valueslist}%
    \!lop\!valueslist\to\!value
    \let\\=\cr\lineskiplimit=\maxdimen\lineskip=#1%
    \baselineskip=-1000pt\halign{\!lglue##\!rglue\cr \!value\!valueslist\cr}}%
  \ignorespaces}

\def\!lines[#1]#2{%
  \let\!lglue=\hfill \let\!rglue=\hfill
  \expandafter\let\csname !#1glue\endcsname=\relax
  \vbox{\halign{\!lglue##\!rglue\cr #2\crcr}}%
  \ignorespaces}

\def\!Lines[#1]#2{%
  \let\!lglue=\hfill \let\!rglue=\hfill
  \expandafter\let\csname !#1glue\endcsname=\relax
  \vtop{\halign{\!lglue##\!rglue\cr #2\crcr}}%
  \ignorespaces}

\def\setplotsymbol(#1#2){%
  \!setputobject{#1}{#2}
  \setbox\!plotsymbol=\box\!putobject%
  \!plotsymbolxshift=\!xshift 
  \!plotsymbolyshift=\!yshift 
  \ignorespaces}
 
\font\fiverm=cmr5
\setplotsymbol({\fiverm .})

\def\!!plot(#1,#2){%
  \!dimenA=-\!plotxorigin \advance \!dimenA by #1
  \!dimenB=-\!plotyorigin \advance \!dimenB by #2
  \kern\!dimenA\raise\!dimenB\copy\!plotsymbol\kern-\!dimenA%
  \ignorespaces}
 
\def\!!!plot(#1,#2){%
  \!dimenA=-\!plotxorigin \advance \!dimenA by #1
  \!dimenB=-\!plotyorigin \advance \!dimenB by #2
  \kern\!dimenA\raise\!dimenB\copy\!plotsymbol\kern-\!dimenA%
  \!countE=\!dimenA
  \!countF=\!dimenB
  \immediate\write\!replotfile{\the\!countE,\the\!countF.}%
  \ignorespaces}

\def\savelinesandcurves on "#1" {%
  \immediate\closeout\!replotfile
  \immediate\openout\!replotfile=#1%
  \let\!plot=\!!!plot}

\def\dontsavelinesandcurves {%
  \let\!plot=\!!plot}
\dontsavelinesandcurves

{\catcode`\%=11\xdef\!Commentsignal{
\def\writesavefile#1 {%
  \immediate\write\!replotfile{\!Commentsignal #1}%
  \ignorespaces}

\def\replot"#1" {%
  \expandafter\!replot\input #1 /}
\def\!replot#1,#2. {%
  \!dimenA=#1sp
  \kern\!dimenA\raise#2sp\copy\!plotsymbol\kern-\!dimenA
  \futurelet\!nextchar\!!replot}
\def\!!replot{%
  \if /\!nextchar 
    \def\!next{\!finish}%
  \else
    \def\!next{\!replot}%
  \fi
  \!next}

\def\!Pythag#1#2#3{%
  \!dimenE=#1\relax                                     
  \ifdim\!dimenE<\!zpt 
    \!dimenE=-\!dimenE 
  \fi
  \!dimenF=#2\relax
  \ifdim\!dimenF<\!zpt 
    \!dimenF=-\!dimenF 
  \fi
  \advance \!dimenF by \!dimenE
  \ifdim\!dimenF=\!zpt 
    \!dimenG=\!zpt
  \else 
    \!divide{8\!dimenE}\!dimenF\!dimenE
    \advance\!dimenE by -4pt
      \!dimenE=2\!dimenE
    \!removept\!dimenE\!!t
    \!dimenE=\!!t\!dimenE
    \advance\!dimenE by 64pt
    \divide \!dimenE by 2
    \!dimenH=7pt
    \!!Pythag\!!Pythag\!!Pythag
    \!removept\!dimenH\!!t
    \!dimenG=\!!t\!dimenF
    \divide\!dimenG by 8
  \fi
  #3=\!dimenG
  \ignorespaces}

\def\!!Pythag{
  \!divide\!dimenE\!dimenH\!dimenI
  \advance\!dimenH by \!dimenI
    \divide\!dimenH by 2}

\def\placehypotenuse for <#1> and <#2> in <#3> {%
  \!Pythag{#1}{#2}{#3}}

\def\!qjoin (#1,#2) (#3,#4){%
  \advance\!intervalno by 1
  \!ifcoordmode
    \edef\!xmidpt{#1}\edef\!ymidpt{#2}%
  \else
    \!dimenA=#1\relax \edef\!xmidpt{\the\!dimenA}%
    \!dimenA=#2\relax \edef\!ymidpt{\the\!dimenA}%
  \fi
  \!xM=\!M{#1}\!xunit  \!yM=\!M{#2}\!yunit   \!rotateaboutpivot\!xM\!yM
  \!xE=\!M{#3}\!xunit  \!yE=\!M{#4}\!yunit   \!rotateaboutpivot\!xE\!yE
  \!dimenA=\!xM  \advance \!dimenA by -\!xS
  \!dimenB=\!xE  \advance \!dimenB by -\!xM
  \!xB=3\!dimenA \advance \!xB by -\!dimenB
  \!xC=2\!dimenB \advance \!xC by -2\!dimenA
  \!dimenA=\!yM  \advance \!dimenA by -\!yS%
  \!dimenB=\!yE  \advance \!dimenB by -\!yM%
  \!yB=3\!dimenA \advance \!yB by -\!dimenB%
  \!yC=2\!dimenB \advance \!yC by -2\!dimenA%
  \!xprime=\!xB  \!yprime=\!yB
  \!dxprime=.5\!xC  \!dyprime=.5\!yC
  \!getf \!midarclength=\!dimenA
  \!getf \advance \!midarclength by 4\!dimenA
  \!getf \advance \!midarclength by \!dimenA
  \divide \!midarclength by 12
  \!arclength=\!dimenA
  \!getf \advance \!arclength by 4\!dimenA
  \!getf \advance \!arclength by \!dimenA
  \divide \!arclength by 12
  \advance \!arclength by \!midarclength
  \global\advance \totalarclength by \!arclength
  \ifdim\!distacross>\!arclength 
    \advance \!distacross by -\!arclength
  \else
    \!initinverseinterp
    \loop\ifdim\!distacross<\!arclength
      \!inverseinterp
      \!xpos=\!t\!xC \advance\!xpos by \!xB
        \!xpos=\!t\!xpos \advance \!xpos by \!xS
      \!ypos=\!t\!yC \advance\!ypos by \!yB
        \!ypos=\!t\!ypos \advance \!ypos by \!yS
      \!plotifinbounds
      \advance\!distacross \plotsymbolspacing
      \!advancedashing
    \repeat  
    \advance \!distacross by -\!arclength
  \fi
  \!xS=\!xE
  \!yS=\!yE
  \ignorespaces}

\def\!getf{\!Pythag\!xprime\!yprime\!dimenA%
  \advance\!xprime by \!dxprime
  \advance\!yprime by \!dyprime}

\def\!initinverseinterp{%
  \ifdim\!arclength>\!zpt
    \!divide{8\!midarclength}\!arclength\!dimenE
    \ifdim\!dimenE<\!wmin \!setinverselinear
    \else 
      \ifdim\!dimenE>\!wmax \!setinverselinear
      \else
        \def\!inverseinterp{\!inversequad}\ignorespaces
         \!removept\!dimenE\!Ew
         \!dimenF=-\!Ew\!dimenE
         \advance\!dimenF by 32pt
         \!dimenG=8pt 
         \advance\!dimenG by -\!dimenE
         \!dimenG=\!Ew\!dimenG
         \!divide\!dimenF\!dimenG\!beta
         \!gamma=1pt
         \advance \!gamma by -\!beta
      \fi
    \fi
  \fi
  \ignorespaces}

\def\!inversequad{%
  \!divide\!distacross\!arclength\!dimenG
  \!removept\!dimenG\!v
  \!dimenG=\!v\!gamma
  \advance\!dimenG by \!beta
  \!dimenG=\!v\!dimenG
  \!removept\!dimenG\!t}

\def\!setinverselinear{%
  \def\!inverseinterp{\!inverselinear}%
  \divide\!dimenE by 8 \!removept\!dimenE\!t
  \!countC=\!intervalno \multiply \!countC 2
  \!countB=\!countC     \advance \!countB -1
  \!countA=\!countB     \advance \!countA -1
  \wlog{\the\!countB th point (\!xmidpt,\!ymidpt) being plotted 
    doesn't lie in the}%
  \wlog{ middle third of the arc between the \the\!countA th 
    and \the\!countC th points:}%
  \wlog{ [arc length \the\!countA\space to \the\!countB]/[arc length 
    \the \!countA\space to \the\!countC]=\!t.}%
  \ignorespaces}
 
\def\!inverselinear{%
  \!divide\!distacross\!arclength\!dimenG
  \!removept\!dimenG\!t}

\def\startrotation{%
  \let\!rotateaboutpivot=\!!rotateaboutpivot
  \let\!rotateonly=\!!rotateonly
  \!ifnextchar{b}{\!getsincos }%
    {\!getsincos by {\!cosrotationangle} {\!sinrotationangle} }}
\def\!getsincos by #1 #2 {%
  \edef\!cosrotationangle{#1}%
  \edef\!sinrotationangle{#2}%
  \!ifcoordmode 
    \let\!ROnext=\!ccheckforpivot
  \else
    \let\!ROnext=\!dcheckforpivot
  \fi
  \!ROnext}
\def\!ccheckforpivot{%
  \!ifnextchar{a}{\!cgetpivot}%
    {\!cgetpivot about {\!xpivotcoord} {\!ypivotcoord} }}
\def\!cgetpivot about #1 #2 {%
  \edef\!xpivotcoord{#1}%
  \edef\!ypivotcoord{#2}%
  \!xpivot=#1\!xunit  \!ypivot=#2\!yunit
  \ignorespaces}
\def\!dcheckforpivot{%
  \!ifnextchar{a}{\!dgetpivot}{\ignorespaces}}
\def\!dgetpivot about #1 #2 {%
  \!xpivot=#1\relax  \!ypivot=#2\relax
  \ignorespaces}

\def\stoprotation{%
  \let\!rotateaboutpivot=\!!!rotateaboutpivot
  \let\!rotateonly=\!!!rotateonly
  \ignorespaces}
 
\def\!!rotateaboutpivot#1#2{%
  \!dimenA=#1\relax  \advance\!dimenA -\!xpivot
  \!dimenB=#2\relax  \advance\!dimenB -\!ypivot
  \!dimenC=\!cosrotationangle\!dimenA
    \advance \!dimenC -\!sinrotationangle\!dimenB
  \!dimenD=\!cosrotationangle\!dimenB
    \advance \!dimenD  \!sinrotationangle\!dimenA
  \advance\!dimenC \!xpivot  \advance\!dimenD \!ypivot
  #1=\!dimenC  #2=\!dimenD
  \ignorespaces}

\def\!!rotateonly#1#2{%
  \!dimenA=#1\relax  \!dimenB=#2\relax 
  \!dimenC=\!cosrotationangle\!dimenA
    \advance \!dimenC -\!rotsign\!sinrotationangle\!dimenB
  \!dimenD=\!cosrotationangle\!dimenB
    \advance \!dimenD  \!rotsign\!sinrotationangle\!dimenA
  #1=\!dimenC  #2=\!dimenD
  \ignorespaces}
\def\!rotsign{}
\def\!!!rotateaboutpivot#1#2{\relax}
\def\!!!rotateonly#1#2{\relax}
\stoprotation

\def\!reverserotateonly#1#2{%
  \def\!rotsign{-}%
  \!rotateonly{#1}{#2}%
  \def\!rotsign{}%
  \ignorespaces}

\def\!getspan span <#1>{%
  \!dshade=#1\relax
  \!ifcoordmode 
    \let\!GRnext=\!GRccheckforAP
  \else
    \let\!GRnext=\!GRdcheckforAP
  \fi
  \!GRnext}
\def\!GRccheckforAP{%
  \!ifnextchar{p}{\!cgetanchor }
    {\!cgetanchor point at {\!xshadesave} {\!yshadesave} }}
\def\!cgetanchor point at #1 #2 {%
  \edef\!xshadesave{#1}\edef\!yshadesave{#2}%
  \!xshade=\!xshadesave\!xunit  \!yshade=\!yshadesave\!yunit
  \ignorespaces}
\def\!GRdcheckforAP{%
  \!ifnextchar{p}{\!dgetanchor}%
    {\ignorespaces}}
\def\!dgetanchor point at #1 #2 {%
  \!xshade=#1\relax  \!yshade=#2\relax
  \ignorespaces}

\def\setshadesymbol{%
  \!ifnextchar<{\!setshadesymbol}{\!setshadesymbol<,,,> }}

\def\!setshadesymbol <#1,#2,#3,#4> (#5#6){%
  \!setputobject{#5}{#6}%
  \setbox\!shadesymbol=\box\!putobject%
  \!shadesymbolxshift=\!xshift \!shadesymbolyshift=\!yshift
  \!dimenA=\!xshift \advance\!dimenA \!smidge
  \!override\!dimenA{#1}\!lshrinkage%
  \!dimenA=\!wd \advance \!dimenA -\!xshift
    \advance\!dimenA \!smidge
    \!override\!dimenA{#2}\!rshrinkage
  \!dimenA=\!dp \advance \!dimenA \!yshift
    \advance\!dimenA \!smidge
    \!override\!dimenA{#3}\!bshrinkage
  \!dimenA=\!ht \advance \!dimenA -\!yshift
    \advance\!dimenA \!smidge
    \!override\!dimenA{#4}\!tshrinkage
  \ignorespaces}
\def\!smidge{-.2pt}%

\def\!override#1#2#3{%
  \edef\!!override{#2}%
  \ifx \!!override\empty
    #3=#1\relax
  \else
    \if z\!!override
      #3=\!zpt
    \else
      \ifx \!!override\!blankz
        #3=\!zpt
      \else
        #3=#2\relax
      \fi
    \fi
  \fi
  \ignorespaces}
\def\!blankz{ z}

\setshadesymbol ({\fiverm .})

\def\!startvshade#1(#2,#3,#4){%
  \let\!!xunit=\!xunit%
  \let\!!yunit=\!yunit%
  \let\!!xshade=\!xshade%
  \let\!!yshade=\!yshade%
  \def\!getshrinkages{\!vgetshrinkages}%
  \let\!setshadelocation=\!vsetshadelocation%
  \!xS=\!M{#2}\!!xunit
  \!ybS=\!M{#3}\!!yunit
  \!ytS=\!M{#4}\!!yunit
  \!shadexorigin=\!xorigin  \advance \!shadexorigin \!shadesymbolxshift
  \!shadeyorigin=\!yorigin  \advance \!shadeyorigin \!shadesymbolyshift
  \ignorespaces}
 
\def\!starthshade#1(#2,#3,#4){%
  \let\!!xunit=\!yunit%
  \let\!!yunit=\!xunit%
  \let\!!xshade=\!yshade%
  \let\!!yshade=\!xshade%
  \def\!getshrinkages{\!hgetshrinkages}%
  \let\!setshadelocation=\!hsetshadelocation%
  \!xS=\!M{#2}\!!xunit
  \!ybS=\!M{#3}\!!yunit
  \!ytS=\!M{#4}\!!yunit
  \!shadexorigin=\!xorigin  \advance \!shadexorigin \!shadesymbolxshift
  \!shadeyorigin=\!yorigin  \advance \!shadeyorigin \!shadesymbolyshift
  \ignorespaces}

\def\!lattice#1#2#3#4#5{%
  \!dimenA=#1
  \!dimenB=#2
  \!countB=\!dimenB
  \!dimenC=#3
  \advance\!dimenC -\!dimenA
  \!countA=\!dimenC
  \divide\!countA \!countB
  \ifdim\!dimenC>\!zpt
    \!dimenD=\!countA\!dimenB
    \ifdim\!dimenD<\!dimenC
      \advance\!countA 1 
    \fi
  \fi
  \!dimenC=\!countA\!dimenB
    \advance\!dimenC \!dimenA
  #4=\!countA
  #5=\!dimenC
  \ignorespaces}

\def\!qshade#1(#2,#3,#4)#5(#6,#7,#8){%
  \!xM=\!M{#2}\!!xunit
  \!ybM=\!M{#3}\!!yunit
  \!ytM=\!M{#4}\!!yunit
  \!xE=\!M{#6}\!!xunit
  \!ybE=\!M{#7}\!!yunit
  \!ytE=\!M{#8}\!!yunit
  \!getcoeffs\!xS\!ybS\!xM\!ybM\!xE\!ybE\!ybB\!ybC
  \!getcoeffs\!xS\!ytS\!xM\!ytM\!xE\!ytE\!ytB\!ytC
  \def\!getylimits{\!qgetylimits}%
  \!shade{#1}\ignorespaces}
 
\def\!lshade#1(#2,#3,#4){%
  \!xE=\!M{#2}\!!xunit
  \!ybE=\!M{#3}\!!yunit
  \!ytE=\!M{#4}\!!yunit
  \!dimenE=\!xE  \advance \!dimenE -\!xS
  \!dimenC=\!ytE \advance \!dimenC -\!ytS
  \!divide\!dimenC\!dimenE\!ytB
  \!dimenC=\!ybE \advance \!dimenC -\!ybS
  \!divide\!dimenC\!dimenE\!ybB
  \def\!getylimits{\!lgetylimits}%
  \!shade{#1}\ignorespaces}
 
\def\!getcoeffs#1#2#3#4#5#6#7#8{%
  \!dimenC=#4\advance \!dimenC -#2
  \!dimenE=#3\advance \!dimenE -#1
  \!divide\!dimenC\!dimenE\!dimenF
  \!dimenC=#6\advance \!dimenC -#4
  \!dimenH=#5\advance \!dimenH -#3
  \!divide\!dimenC\!dimenH\!dimenG
  \advance\!dimenG -\!dimenF
  \advance \!dimenH \!dimenE
  \!divide\!dimenG\!dimenH#8
  \!removept#8\!t
  #7=-\!t\!dimenE
  \advance #7\!dimenF
  \ignorespaces}

\def\!shade#1{%
  \!getshrinkages#1<,,,>\!nil
  \advance \!dimenE \!xS
  \!lattice\!!xshade\!dshade\!dimenE
    \!parity\!xpos
  \!dimenF=-\!dimenF
    \advance\!dimenF \!xE
  \!loop\!not{\ifdim\!xpos>\!dimenF}
    \!shadecolumn%
    \advance\!xpos \!dshade
    \advance\!parity 1
  \repeat
  \!xS=\!xE
  \!ybS=\!ybE
  \!ytS=\!ytE
  \ignorespaces}

\def\!vgetshrinkages#1<#2,#3,#4,#5>#6\!nil{%
  \!override\!lshrinkage{#2}\!dimenE
  \!override\!rshrinkage{#3}\!dimenF
  \!override\!bshrinkage{#4}\!dimenG
  \!override\!tshrinkage{#5}\!dimenH
  \ignorespaces}
\def\!hgetshrinkages#1<#2,#3,#4,#5>#6\!nil{%
  \!override\!lshrinkage{#2}\!dimenG
  \!override\!rshrinkage{#3}\!dimenH
  \!override\!bshrinkage{#4}\!dimenE
  \!override\!tshrinkage{#5}\!dimenF
  \ignorespaces}

\def\!shadecolumn{%
  \!dxpos=\!xpos
  \advance\!dxpos -\!xS
  \!removept\!dxpos\!dx
  \!getylimits
  \advance\!ytpos -\!dimenH
  \advance\!ybpos \!dimenG
  \!yloc=\!!yshade
  \ifodd\!parity 
     \advance\!yloc \!dshade
  \fi
  \!lattice\!yloc{2\!dshade}\!ybpos%
    \!countA\!ypos
  \!dimenA=-\!shadexorigin \advance \!dimenA \!xpos
  \loop\!not{\ifdim\!ypos>\!ytpos}
    \!setshadelocation
    \!rotateaboutpivot\!xloc\!yloc%
    \!dimenA=-\!shadexorigin \advance \!dimenA \!xloc
    \!dimenB=-\!shadeyorigin \advance \!dimenB \!yloc
    \kern\!dimenA \raise\!dimenB\copy\!shadesymbol \kern-\!dimenA
    \advance\!ypos 2\!dshade
  \repeat
  \ignorespaces}
 
\def\!qgetylimits{%
  \!dimenA=\!dx\!ytC              
  \advance\!dimenA \!ytB
  \!ytpos=\!dx\!dimenA
  \advance\!ytpos \!ytS
  \!dimenA=\!dx\!ybC              
  \advance\!dimenA \!ybB
  \!ybpos=\!dx\!dimenA
  \advance\!ybpos \!ybS}
 
\def\!lgetylimits{%
  \!ytpos=\!dx\!ytB
  \advance\!ytpos \!ytS
  \!ybpos=\!dx\!ybB
  \advance\!ybpos \!ybS}
 
\def\!vsetshadelocation{
  \!xloc=\!xpos
  \!yloc=\!ypos}
\def\!hsetshadelocation{
  \!xloc=\!ypos
  \!yloc=\!xpos}

\def\!axisticks {%
  \def\!nextkeyword##1 {%
    \expandafter\ifx\csname !ticks##1\endcsname \relax
      \def\!next{\!fixkeyword{##1}}%
    \else
      \def\!next{\csname !ticks##1\endcsname}%
    \fi
    \!next}%
  \!axissetup
    \def\!axissetup{\relax}%
  \edef\!ticksinoutsign{\!ticksinoutSign}%
  \!ticklength=\longticklength
  \!tickwidth=\linethickness
  \!gridlinestatus
  \!setticktransform
  \!maketick
  \!tickcase=0
  \def\!LTlist{}%
  \!nextkeyword}

\def\ticksout{%
  \def\!ticksinoutSign{+}}

\ticksout

\def\nogridlines{%
  \def\!gridlinestatus{\!gridlinestoofalse}}
\nogridlines

\def\loggedticks{%
  \def\!setticktransform{\let\!ticktransform=\!logten}}
\def\unloggedticks{%
  \def\!setticktransform{\let\!ticktransform=\!donothing}}
\def\!donothing#1#2{\def#2{#1}}
\unloggedticks

\expandafter\def\csname !ticks/\endcsname{%
  \!not {\ifx \!LTlist\empty}
    \!placetickvalues
  \fi
  \def\!tickvalueslist{}%
  \def\!LTlist{}%
  \expandafter\csname !axis/\endcsname}

\def\!maketick{%
  \setbox\!boxA=\hbox{%
    \beginpicture
      \!setdimenmode
      \setcoordinatesystem point at {\!zpt} {\!zpt}   
      \linethickness=\!tickwidth
      \ifdim\!ticklength>\!zpt
        \putrule from {\!zpt} {\!zpt} to
          {\!ticksinoutsign\!tickxsign\!ticklength}
          {\!ticksinoutsign\!tickysign\!ticklength}
      \fi
      \if!gridlinestoo
        \putrule from {\!zpt} {\!zpt} to
          {-\!tickxsign\!xaxislength} {-\!tickysign\!yaxislength}
      \fi
    \endpicturesave <\!Xsave,\!Ysave>}%
    \wd\!boxA=\!zpt}
  
\def\!ticksin{%
  \def\!ticksinoutsign{-}%
  \!maketick
  \!nextkeyword}

\def\!ticksout{%
  \def\!ticksinoutsign{+}%
  \!maketick
  \!nextkeyword}

\def\!tickslength<#1> {%
  \!ticklength=#1\relax
  \!maketick
  \!nextkeyword}

\def\!tickslong{%
  \!tickslength<\longticklength> }

\def\!ticksshort{%
  \!tickslength<\shortticklength> }

\def\!tickswidth<#1> {%
  \!tickwidth=#1\relax
  \!maketick
  \!nextkeyword}

\def\!ticksandacross{%
  \!gridlinestootrue
  \!maketick
  \!nextkeyword}

\def\!ticksbutnotacross{%
  \!gridlinestoofalse
  \!maketick
  \!nextkeyword}

\def\!tickslogged{%
  \let\!ticktransform=\!logten
  \!nextkeyword}

\def\!ticksunlogged{%
  \let\!ticktransform=\!donothing
  \!nextkeyword}

\def\!ticksunlabeled{%
  \!tickcase=0
  \!nextkeyword}

\def\!ticksnumbered{%
  \!tickcase=1
  \!nextkeyword}

\def\!tickswithvalues#1/ {%
  \edef\!tickvalueslist{#1! /}%
  \!tickcase=2
  \!nextkeyword}

\def\!ticksquantity#1 {%
  \ifnum #1>1
    \!updatetickoffset
    \!countA=#1\relax
    \advance \!countA -1
    \!ticklocationincr=\!axisLength
      \divide \!ticklocationincr \!countA
    \!ticklocation=\!axisstart
    \loop \!not{\ifdim \!ticklocation>\!axisend}
      \!placetick\!ticklocation
      \ifcase\!tickcase
          \relax 
        \or
          \relax 
        \or
          \expandafter\!gettickvaluefrom\!tickvalueslist
          \edef\!tickfield{{\the\!ticklocation}{\!value}}%
          \expandafter\!listaddon\expandafter{\!tickfield}\!LTlist%
      \fi
      \advance \!ticklocation \!ticklocationincr
    \repeat
  \fi
  \!nextkeyword}

\def\!ticksat#1 {%
  \!updatetickoffset
  \edef\!Loc{#1}%
  \if /\!Loc
    \def\next{\!nextkeyword}%
  \else
    \!ticksincommon
    \def\next{\!ticksat}%
  \fi
  \next}    
      
\def\!ticksfrom#1 to #2 by #3 {%
  \!updatetickoffset
  \edef\!arg{#3}%
  \expandafter\!separate\!arg\!nil
  \!scalefactor=1
  \expandafter\!countfigures\!arg/
  \edef\!arg{#1}%
  \!scaleup\!arg by\!scalefactor to\!countE
  \edef\!arg{#2}%
  \!scaleup\!arg by\!scalefactor to\!countF
  \edef\!arg{#3}%
  \!scaleup\!arg by\!scalefactor to\!countG
  \loop \!not{\ifnum\!countE>\!countF}
    \ifnum\!scalefactor=1
      \edef\!Loc{\the\!countE}%
    \else
      \!scaledown\!countE by\!scalefactor to\!Loc
    \fi
    \!ticksincommon
    \advance \!countE \!countG
  \repeat
  \!nextkeyword}

\def\!updatetickoffset{%
  \!dimenA=\!ticksinoutsign\!ticklength
  \ifdim \!dimenA>\!offset
    \!offset=\!dimenA
  \fi}

\def\!placetick#1{%
  \if!xswitch
    \!xpos=#1\relax
    \!ypos=\!axisylevel
  \else
    \!xpos=\!axisxlevel
    \!ypos=#1\relax
  \fi
  \advance\!xpos \!Xsave
  \advance\!ypos \!Ysave
  \kern\!xpos\raise\!ypos\copy\!boxA\kern-\!xpos
  \ignorespaces}

\def\!gettickvaluefrom#1 #2 /{%
  \edef\!value{#1}%
  \edef\!tickvalueslist{#2 /}%
  \ifx \!tickvalueslist\!endtickvaluelist
    \!tickcase=0
  \fi}
\def\!endtickvaluelist{! /}

\def\!ticksincommon{%
  \!ticktransform\!Loc\!t
  \!ticklocation=\!t\!!unit
  \advance\!ticklocation -\!!origin
  \!placetick\!ticklocation
  \ifcase\!tickcase
    \relax 
  \or 
    \ifdim\!ticklocation<-\!!origin
      \edef\!Loc{$\!Loc$}%
    \fi
    \edef\!tickfield{{\the\!ticklocation}{\!Loc}}%
    \expandafter\!listaddon\expandafter{\!tickfield}\!LTlist%
  \or 
    \expandafter\!gettickvaluefrom\!tickvalueslist
    \edef\!tickfield{{\the\!ticklocation}{\!value}}%
    \expandafter\!listaddon\expandafter{\!tickfield}\!LTlist%
  \fi}

\def\!separate#1\!nil{%
  \!ifnextchar{-}{\!!separate}{\!!!separate}#1\!nil}
\def\!!separate-#1\!nil{%
  \def\!sign{-}%
  \!!!!separate#1..\!nil}
\def\!!!separate#1\!nil{%
  \def\!sign{+}%
  \!!!!separate#1..\!nil}
\def\!!!!separate#1.#2.#3\!nil{%
  \def\!arg{#1}%
  \ifx\!arg\!empty
    \!countA=0
  \else
    \!countA=\!arg
  \fi
  \def\!arg{#2}%
  \ifx\!arg\!empty
    \!countB=0
  \else
    \!countB=\!arg
  \fi}
 
\def\!countfigures#1{%
  \if #1/%
    \def\!next{\ignorespaces}%
  \else
    \multiply\!scalefactor 10
    \def\!next{\!countfigures}%
  \fi
  \!next}

\def\!scaleup#1by#2to#3{%
  \expandafter\!separate#1\!nil
  \multiply\!countA #2\relax
  \advance\!countA \!countB
  \if -\!sign
    \!countA=-\!countA
  \fi
  #3=\!countA
  \ignorespaces}

\def\!scaledown#1by#2to#3{%
  \!countA=#1\relax
  \ifnum \!countA<0 
    \def\!sign{-}
    \!countA=-\!countA
  \else
    \def\!sign{}%
  \fi
  \!countB=\!countA
  \divide\!countB #2\relax
  \!countC=\!countB
    \multiply\!countC #2\relax
  \advance \!countA -\!countC
  \edef#3{\!sign\the\!countB.}
  \!countC=\!countA 
  \ifnum\!countC=0 
    \!countC=1
  \fi
  \multiply\!countC 10
  \!loop \ifnum #2>\!countC
    \edef#3{#3\!zero}%
    \multiply\!countC 10
  \repeat
  \edef#3{#3\the\!countA}
  \ignorespaces}

\def\!placetickvalues{%
  \advance\!offset \tickstovaluesleading
  \if!xswitch
    \setbox\!boxA=\hbox{%
      \def\\##1##2{%
        \!dimenput {##2} [B] (##1,\!axisylevel)}%
      \beginpicture 
        \!LTlist
      \endpicturesave <\!Xsave,\!Ysave>}%
    \!dimenA=\!axisylevel
      \advance\!dimenA -\!Ysave
      \advance\!dimenA \!tickysign\!offset
      \if -\!tickysign
        \advance\!dimenA -\ht\!boxA
      \else
        \advance\!dimenA  \dp\!boxA
      \fi
    \advance\!offset \ht\!boxA 
      \advance\!offset \dp\!boxA
    \!dimenput {\box\!boxA} [Bl] <\!Xsave,\!Ysave> (\!zpt,\!dimenA)
  \else
    \setbox\!boxA=\hbox{%
      \def\\##1##2{%
        \!dimenput {##2} [r] (\!axisxlevel,##1)}%
      \beginpicture 
        \!LTlist
      \endpicturesave <\!Xsave,\!Ysave>}%
    \!dimenA=\!axisxlevel
      \advance\!dimenA -\!Xsave
      \advance\!dimenA \!tickxsign\!offset
      \if -\!tickxsign
        \advance\!dimenA -\wd\!boxA
      \fi
    \advance\!offset \wd\!boxA
    \!dimenput {\box\!boxA} [Bl] <\!Xsave,\!Ysave> (\!dimenA,\!zpt)
  \fi}

\normalgraphs
\catcode`!=12 

\sloppy

\begin{document}


\renewcommand{\thefootnote}{\fnsymbol{footnote}}
\renewcommand{\thefootnote}{\arabic{footnote}}


\title{\bf A Morphology-System and\\
Part-of-Speech Tagger for German\footnote{In:
D. Gibbon, ed., \it Natural Language Processing and Speech Technology.
Results of the 3rd KONVENS Conference, Bielefeld, October 1996. \rm Mouton de Gruyter, Berlin, 1996.}}

\author{Wolfgang Lezius, Reinhard Rapp \& Manfred Wettler}

\date{}

\maketitle

\thispagestyle{empty}


\vspace{-2ex}

\noindent This paper presents an integrated tool for German morphology and
statistical part-of-speech tagging which aims at making some well established
methods widely available. The software is very user friendly, runs on any PC and
can be downloaded as a complete package (including lexicon and documentation) from
the World Wide Web. Compared with the performance of other tagging systems the
tagger produces similar results.

\vspace{1.25ex}

\noindent Es wird ein integriertes Programmpaket vorgestellt, das ein
Morphologie- und ein Taggingmodul f{\"u}r das Deutsche enth{\"alt}.
Die frei verf{\"ug}bare Software zeichnet sich insbesondere durch hohe
Benutzerfreundlichkeit aus und kann {\"u}ber das World Wide Web bezogen werden.
Die Qualit{\"a}t der erzielten Ergebnisse entspricht dem derzeitigen Stand
der Forschung.

\vspace{-0.25ex}


\setlength{\parindent}{0em}
\setlength{\parskip}{1ex}


\section{Introduction}

Morphology systems, lemmatisers and part-of-speech taggers are some of the
basic tools in natural language processing. There are numerous applications,
including syntax parsing, machine translation, automatic indexing and semantic
clustering of words. Unfortunately, for languages other than English, such tools
are rarely available, and different research groups are often forced to
redevelop them over and over again. Considering German, quite a few morphology
systems (Hausser 1996) and taggers (see table \ref{Tagger}) have been developed, which are described in Wothke et al. (1993) (IBM Heidelberg), Steiner (1995) (University of M{\"u}nster), Feldweg (1995) (University of T{\"u}bingen), Schmid (1995) (University of Stuttgart), Armstrong et al. (1995) (ISSCO Geneva), and Lezius (1995) (University of Paderborn).
However, in most cases, the tagger is isolated from the morphology system. It relies on a lexicon of full forms which, of course, may be generated by a morphological
tool. Unfortunately, most German lexicons are not available due to copyright
restrictions and - as far as we know - none of them is public-domain.
Therefore we have decided to make our system Morphy publicly available.
It combines a morphological and tagging module in a single package
and can be downloaded from the World Wide
Web.\footnote{
URL: http://www-psycho.uni-paderborn.de/lezius/morpho.html}

\begin{table}
\caption{\label{Tagger} Comparison of German Taggers}

\begin{footnotesize}

\begin{center}
\begin{tabular}{|l||c|c|c|c|c|c|} \hline
 & IBM & Univer- & Univer- & Univer- & & Univer- \\
 & Heidel- & sity of & sity of & sity of & ISSCO & sity of \\
 & berg & M{\"u}nster & T{\"u}bingen & Stuttgart & Geneva & Paderborn \\
\hline \hline

learning & super- & super- & unsuper- & unsuper- & unsuper- & super-\\
method & vised & vised & vised & vised & vised & vised \\ \hline

context & bi- \& tri- & bi- & bi- & bi- & bi- & tri- \\
method & grams & grams & grams & grams & grams & grams \\ \hline

training & 20.000 & 200.000 & 200.000 & 20.000 & 20.000 & 20.000 \\
corpus & words & words & words & words & words & words \\ \hline

test & 10.000 & 30.000 & 20.000 & 5.000 & 1.850 & 5.000 \\
corpus & words & words & words & words & words & words \\ \hline

size of & 534.514 & 850.000 & 500.000 & 350.000 & 30.000 & 100.000 \\
lexicon & words & words & words & words & words & words \\ \hline

tag sets & 689 & 143 & -  & - & - & 456 \\
large / small & 33 & 54 & 42 & 50 & 56 & 51 \\ \hline

accuracy & 77.7\% & 81.5\% & - & - & - & 84.7\% \\
large / small & 93.4\% & 92.8\% & 96.7\% & 97.0\% & 96.5\% & 95.9\% \\ \hline

\end{tabular}
\end{center}

\end{footnotesize}

\end{table}

Since it has been created not only for linguists, but also for second language
learners, it has been designed for standard PC-platforms and great effort has been put in making it as easy to use as possible.

\vspace{-1ex}


\section{The morphology module of Morphy}

The morphology system is based on the Duden grammar (Drosdowski 1984).
It consists of three parts: {\it Analysis}, {\it generation} and
{\it lexical system}.

The lexical system is more sophisticated than other systems in order to allow a
user-friendly extension of the lexicon.
When entering a new word,
the user is asked the minimal number of questions
which are necessary to infer the new word's grammatical features and
which any native speaker should
be able to answer. In most cases only the root of the word has to be typed in,
questions are answered by pressing the number of the correct alternative
(see figure \ref{Lex} for the dialogue when entering the verb
{\it telefonieren}). Currently,
the lexicon comprises 21.500 words (about 100.000 word forms)
and is extended continuously.

Starting from the root of a word and the inflexion type as stored in the
lexicon, the generation system produces all inflected
forms which are shown on the screen.
Among other morphological features it considers vowel mutation, shifts
between {\it {\ss}} and {\it ss} as well as pre- and infixation
of markers for participles.

The analysis system for each word form determines its root and its part of speech,
and, if appropiate, its gender, case, number, tense and comparative
degree. It also segments compound nouns using a longest-matching rule
which works from right to left. Since the system treats each word form separately,
ambiguities can not be resolved. For ambiguous word forms any possible lemma and
morphological description is given (for some examples
see table \ref{analysis}). If a word form can not be recognised,
its part of speech is predicted by an algorithm which makes use of
statistical data on German suffix frequencies.

\begin{figure}

\begin{center}
\begin{tabular}{|l|} \hline\\
\hspace*{1.2em}1. Geben Sie den Stamm ein: {\bf telefonieren} \\
\hspace*{1.2em}2. Wird das Verb schwach konjugiert?\\
\hspace*{2.7em}{\bf 1}: Ja\\
\hspace*{2.7em}2: Nein\\
\hspace*{1.2em}3. Wie lautet die 2. Person Singular Pr{\"a}sens? \hspace*{1.2em}\\
\hspace*{2.7em}{\bf 1}: du telefonierst\\
\hspace*{2.7em}2: du telefonierest\\
\hspace*{2.7em}3: du telefoniert\\
\hspace*{1.2em}4. Wie lautet das Partizip des Verbs?\\
\hspace*{2.7em}{\bf 1}: telefoniert\\
\hspace*{2.7em}2: getelefoniert\\
\hspace*{1.2em}Verb klassifiziert!\\
\\ \hline
\end{tabular}
\end{center}

\caption{\label{Lex} Dialogue when entering {\it telefonieren}
(user input is printed bold type)}

\end{figure}

Morphy's lookup-mechanism when analyzing texts is not based on a lexicon of
full forms. Instead, there is only a lexicon of roots together with their
inflexion types. When analyzing a word form, Morphy cuts off all
possible suffixes, builds the possible roots, looks up these roots in the lexicon,
and for each root generates all possible inflected forms. Only those roots
which lead to inflected forms identical to the original word
form will be selected (for details see Lezius 1994).

Naturally, this procedure is much slower than the simple lookup-mechanism
in a full form lexicon.\footnote{Morphy's current analysis speed is about 50 word
forms per second on a fast PC, which is sufficient for many purposes. For the
processing of larger corpora we have
used Morphy to generate a full-form lexicon under UNIX. This has led to an
analysis speed of many thousand word forms per second.}
Nevertheless, there are advantages: First, the lexicon
can be kept small,\footnote{Only 750 KB memory is
necessary for the current lexicon.} which is an important consideration
for a PC-based system intended to be widely distributed. Secondly, the
processing of German compound nouns fits in this concept.

The performance of the morphology system has been tested at
the Morpholympics conference 1994
in Erlangen (see Hausser (1996), pp. 13-14, and Lezius (1996)) with a
specially designed test corpus
which had been unknown to the participants. This corpus comprised about 7.800
word forms and consisted of different text types (two political speeches, a fragment of
the Limas-corpus and a list of special word forms). Morphy
recognised 89.2\%, 95.9\%, 86.9\% and 75.8\% of the word forms, respectively.

\begin{table}
\caption{\label{analysis} Some examples of the morphological analysis}

\begin{footnotesize}

\begin{center}
\begin{tabular}{|l|l|l|} \hline

word form & morphological description & root \\ \hline \hline

Fl{\"u}ssen & Substantiv Dativ Plural maskulinum & Flu{\ss} \\ \hline

Bauern- & Substantiv Dativ Plural neutrum & Bauer / Haus \\
h{\"a}usern & & \\ \hline

Schiffahrts- & Substantiv Genitiv Singular maskulinum & Schiff / Fahrt / \\
hafenmeisters &                                       & Hafen / Meister \\
\hline

K{\"u}sse & Substantiv Nominativ Plural maskulinum & Ku{\ss} \\
       & Substantiv Genitiv Plural maskulinum & Ku{\ss} \\
       & Substantiv Akkusativ Plural maskulinum & Ku{\ss} \\
       & Verb 1. Person Singular Pr{\"a}sens & k{\"u}ssen \\
       & Verb 1. Person Singular Konjunktiv 1 & k{\"u}ssen \\
       & Verb 3. Person Singular Konjunktiv 1 & k{\"u}ssen \\
       & Verb Imperativ Singular & k{\"u}ssen \\ \hline
einnahm & Verb 1. Person Singular Pr{\"a}teritum & (ein)nehmen \\
        & Verb 3. Person Singular Pr{\"a}teritum & (ein)nehmen \\ \hline
verspieltest & Verb 2. Person Singular Pr{\"a}teritum & ver-spielen \\
             & Verb 2. Person Singular Konjunktiv 2 & ver-spielen \\ \hline
verspieltes & Adjektiv Nominativ Singular neutrum & verspielt (ver-spielen)\\
            & Adjektiv Akkusativ Singular neutrum & verspielt (ver-spielen)\\ \hline
edlem & Adjektiv Dativ Singular neutrum & edel \\
      & Adjektiv Dativ Singular maskulinum & edel \\ \hline
\end{tabular}
\end{center}

\end{footnotesize}

\end{table}


\section{The tagging module of Morphy}

Since morphological analysis operates on isolated word forms, ambiguities
are not resolved. The task of the tagger is to resolve these ambiguities
by taking into account contextual information. When designing a tagger,
a number of decisions have to be made:

\begin{itemize}
\item Selection of a tag set.
\item Selection of a tagging algorithm.
\item Selection of a training and test corpus.
\end{itemize}

\subsection{Tag Set}

Like the morphology system, the tagger is
based on the classification of the parts of
speech from the Duden grammar. Supplementary additions have been taken from
the system of Bergenholtz and Schaeder (1977). The so-formed tag
set includes grammatical features as gender,
case and number. This results in a very complex system, comprising about
1000 different tags (see Lezius 1995). Since only 456 tags were actually
used in the training corpus, the tag set was reduced to half.
However, most German word forms are highly
ambiguous in this system (about 5 tags per word form on average).

Although the amount of information gained by this system is very high,
previous tagging algorithms with such large tag sets led to poor results in the
past (see Wothke et al. 1993; Steiner 1995). This is because different grammatical features
often have the same surface realization (e.g. nominative noun and accusative
noun are difficult to distinguish by the tagger). By grouping together parts of
speech with different grammatical features this kind of error can be
significantly reduced. This is what current small tag sets implicitly do.
However, one has to keep in mind that the gain of information provided by the tagger
is also reduced with a smaller tag set.

Since some applications do not require detailed distinctions, we also constructed
a small tag set comprising 51 tags as shown in table \ref{tagset}.
Both tag sets are constructed in such a way
that the large tag set can be directly mapped onto the small tag set.

\begin{table}

\caption{\label{tagset} The small tag set (51 tags)}

\begin{footnotesize}
\begin{center}
\begin{tabular}{||l|l|l||}\hline\hline

tag name & explanation of the tag & example\\ \hline\hline

\tt SUB & Substantiv & \it (der) Mann\\
\tt EIG & Eigenname & \it Egon, (Herr) Hansen\\ \hline

\tt VER & finite Verbform & \it spielst, l{\"a}uft\\
\tt VER INF & Infinitiv & \it spielen, laufen\\
\tt VER PA2 & Partizip Perfekt & \it gespielt, gelaufen\\
\tt VER EIZ & erweiterter Infinitiv mit \it zu & \it abzuspielen\\
\tt VER IMP & Imperativ & \it lauf', laufe\\
\tt VER AUX & finite Hilfsverbform & \it bin, hast\\
\tt VER AUX INF & Infinitiv & \it haben, sein\\
\tt VER AUX PA2 & Partizip Perfekt & \it gahabt, gewesen\\
\tt VER AUX IMP & Imperativ & \it sei, habe\\
\tt VER MOD & finite Modalverbform & \it kannst, will\\
\tt VER MOD INF & Infinitiv & \it k{\"o}nnen, wollen\\
\tt VER MOD PA2 & Partizip Perfekt & \it gekonnt, gewollt\\
\tt VER MOD IMP & Imperativ & \it k{\"o}nne\\ \hline

\tt ART IND & unbestimmter Artikel & \it ein, eines\\
\tt ART DEF & bestimmter Artikel & \it der, des\\ \hline

\tt ADJ & Adjektivform & \it schnelle, kleinstes\\
\tt ADJ ADV & Adjektiv, adverbiell & \it (Er l{\"a}uft) schnell.\\ \hline

\tt PRO DEM ATT & Demonstrativpronomen, attributiv & \it diese
(Frau)\\
\tt PRO DEM PRO & Demonstrativpronomen, pronominal & \it diese\\
\tt PRO REL ATT & Relativpronomen, attributiv & \it ,dessen
(Frau)\\
\tt PRO REL PRO & Relativpronomen, pronominal & \it ,welcher\\
\tt PRO POS ATT & Possesivpronomen, attributiv & \it mein (Buch)\\
\tt PRO POS PRO & Possesivpronomen, pronominal & \it (Das ist)
meiner.\\
\tt PRO IND ATT & Indefinitpronomen, attributiv & \it alle
(Menschen)\\
\tt PRO IND PRO & Indefinitpronomen, pronominal & \it (Ich mag)
alle.\\
\tt PRO INR ATT & Interrogativpronomen, attributiv & \it Welcher
(Mann)?\\
\tt PRO INR PRO & Interrogativpronomen, pronominal & \it Wer?\\
\tt PRO PER & Personalpronomen & \it er, wir\\
\tt PRO REF & Reflexivpronomen & \it sich, uns\\ \hline

\tt ADV & Adverb & \it schon, manchmal\\
\tt ADV PRO & Pronominaladverb & \it damit, dadurch\\
\tt KON UNT & unterordnende Konjunktion & \it da{\ss}, da\\
\tt KON NEB & nebenordnende Konjunktion & \it und, oder\\
\tt KON INF & Infinitivkonjunktion & \it um (zu spielen)\\
\tt KON VGL & Vergleichskonjunktion & \it als, denn, wie\\
\tt KON PRI & Proportionalkonjunktion & \it desto, um so, je\\
\tt PRP & Pr{\"a}position & \it durch, an\\ \hline

\tt SKZ & Sonderklasse f{\"u}r \it zu & \it (,um) zu (spielen)\\
\tt ZUS & Verbzusatz & \it (spielst) ab\\
\tt INJ & Interjektion & \it Wau, Oh\\
\tt ZAL & Zahlw{\"o}rter & \it eins, tausend\\
\tt ZAN & Zahlen & \it 100, 2\\
\tt ABK & Abk{\"u}rzung & \it Dr., usw.\\ \hline

\tt SZD & Doppelpunkt & :\\
\tt SZE & Satzendezeichen & .!?\\
\tt SZG & Gedankenstrich & -\\
\tt SZK & Komma & ,\\
\tt SZS & Semikolon & ;\\
\tt SZN & sonstige Satzzeichen & ()/ \\ \hline \hline

\end{tabular}
\end{center}
\end{footnotesize}

\end{table}

\subsection{Tagging algorithm}

The tagger uses the Church-trigram-algorithm (Church 1988), which is still
unsurpassed in terms of simplicity, robustness and accuracy. However, since
we assumed that longer n-grams may give more information, and
since we observed that some longer n-grams are rather frequent in corpora
(see figure 2 for some statistics on the Brown-corpus),
we decided to compare
the Church algorithm with a tagging algorithm relying on variable context
widths as described by Rapp (1995).

Starting from an ambiguous word form which is to be tagged, this algorithm
considers the preceding word froms - which have already been
tagged - and the succeeding word forms still to be tagged.
For this ambiguous word form the algorithm constructs all possible tag
sequences composed of the already computed tags on the left, one of the
possible tags of the critical word form and possible tags on the right.

The choice of the tag for the critical word form is a function for the
length of the tag sequences to the left and to the right which can be
found in the training corpus. A detailed description of this algorithm
is given in Rapp (Rapp 1995, pp. 149-154).

Although some authors (Cutting et al. 1992; Schmid 1995; Feldweg 1995) claim
that unsupervised tagging algorithms produce superior results, we chose
supervised learning. These publications pay little attention to
the fact that algorithms for unsupervised tagging require great care
(or even luck) when tuning some initial parameters.
It frequently happens that unsupervised learning with sophisticated tag sets
ends up in local minima, which can lead to poor results without any
indication to the user. Such behavior seemed unacceptable for a standard
tool.

\begin{figure}
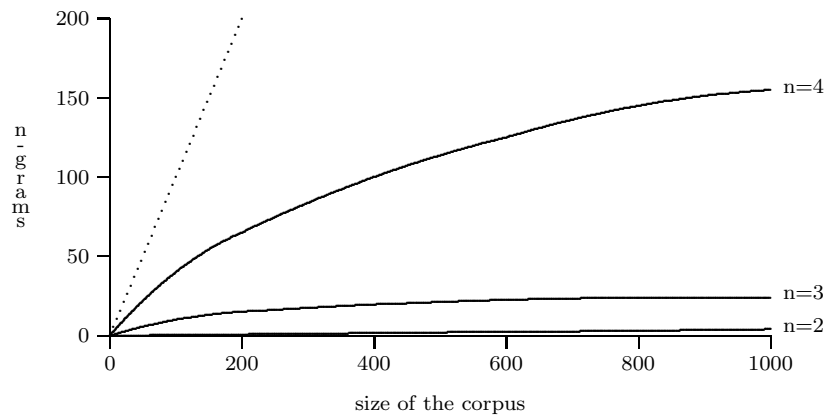


$$\beginpicture

\setcoordinatesystem units <0.025em,0.06em>
\setplotarea x from 0 to 1000, y from 0 to 200

\begin{footnotesize}

\axis bottom label {size of the corpus}
      ticks numbered from 0 to 1000 by 200 /

\axis left label {\stack{n,-,g,r,a,m,s}}
      ticks numbered from 0 to 200 by 50 /

\setplotsymbol ({\rm.})
\linethickness0.01em
\setquadratic

\plotsymbolspacing0.05em


\plot 0 0          100 40      200 65
      400 100      600 125     800 145
      1000 155 /


\plot 0 0          100 10      200 15
      600 22.5     1000 23.5 /


\plot 0 0          500 2       1000 4 /

\setlinear
\plotsymbolspacing0.4em


\plot 0 0  200 200 /

\put {n=2} at 1050 7
\put {n=3} at 1050 27
\put {n=4} at 1050 157.5

\end{footnotesize}
\endpicture$$

\vspace{4ex}

\caption{\label{Brown} Statistics on the Brown corpus: number of different
n-grams occuring in the corpus versus size of the corpus (all figures
in thousands)}

\end{figure}

\subsection{Training and test corpus}

For training and testing we took a fragment from the
``Frankfurter-Rundschau''-corpus,\footnote{This corpus was generously donated by
the Druck- und Verlagshaus Frankfurt
am Main and has been included in the CD-ROM of the European
Corpus Initiative. We thank Gisela Zunker for her help with the acquisition and
preparation of the corpus.}
which we have been collecting since 1992. Tables and other non-textual items
were removed manually. A segment of 20.000 word forms was used for training,
another segment of 5.000 word forms for testing. Any word forms not recognised
by the morphology system
were included in the lexicon. Using a special tagging editor which
- on the basis of the morphology module - for each word gives a choice of
possible tags, both corpora were tagged
semiautomatically with the large tag set. A recent version of the editor
additionally predicts the correct tag.


\section{Results}

Using the probabilities from the manually annotated training corpus,
the test corpus was tagged automatically. The results were compared with the
previous manual annotation of the test corpus. This was done for both tagging
algorithms and tag sets. For the small tag set, the Church algorithm achieved
an accuracy of 95.9\%, whereas with the variable-context algorithm an accuracy
of 95.0\% was obtained. For the large tag set the respective figures are 84.7\%
and 81.8\%.

In comparison with other research groups, the results are
similar for the small tagset and slightly better for the large tagset (see
table \ref{Tagger}). Surprisingly, inspite of considering less context, the Church
algorithm performs better than the variable-context algorithm in both cases.

This is the reason why the current implementation
of Morphy only includes the Church algorithm.\footnote{The speed of the tagger
(including morphological analysis) is about 20 word forms per second for
the large and 100 word forms per second for the small tag set on a fast PC.}
As an example, figure \ref{Example}
gives the annotation results of a few test sentences for both tag sets.

\begin{figure}
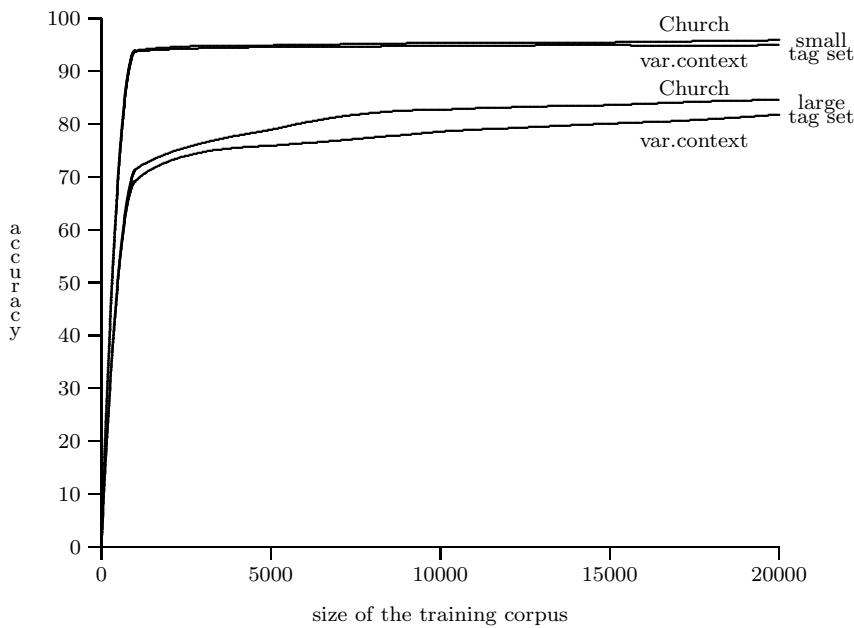


$$\beginpicture

\setcoordinatesystem units <0.001275em,0.2em>
\setplotarea x from 0 to 20000, y from 0 to 100

\begin{footnotesize}

\axis bottom label {size of the training corpus}
      ticks numbered from 0 to 20000 by 5000 /

\axis left label {\stack{a,c,c,u,r,a,c,y}}
      ticks numbered from 0 to 100 by 10 /

\setplotsymbol ({\rm.})
\setquadratic


\plotsymbolspacing0.05em

\plot 0 0          500 70      1000 93.8
      2500 94.3    5000 94.6   7500 94.7
      10000 94.8   12500 94.9  15000 95.0
      17500 94.8   20000 95.0 /


\plotsymbolspacing0.05em

\plot 0 0          500 70      1000 93.9
      2500 94.7    5000 94.9   7500 95.2
      10000 95.4   12500 95.5  15000 95.5
      17500 95.7   20000 95.9 /

\put{Church} at 17500 99
\put{var.context} at 17500 92

\put{small} at 21250 96
\put{tag set} at 21250 93

\setplotsymbol ({\rm.})
\setquadratic


\plotsymbolspacing0.05em

\plot 0 0          500 51      1000 69.2
      2500 74.0    5000 75.9   7500 77.2
      10000 78.6   12500 79.4  15000 80.1
      17500 80.8   20000 81.8 /


\plotsymbolspacing0.05em

\plot 0 0          500 51      1000 71.4
      2500 75.6    5000 78.9   7500 81.8
      10000 82.7   12500 83.3  15000 83.6
      17500 84.2   20000 84.7 /

\put{Church} at 17500 87
\put{var.context} at 17500 77

\put{large} at 21250 84
\put{tag set} at 21250 81

\end{footnotesize}
\endpicture$$

\vspace{4ex}

\caption{\label{Results} Accuracy versus size of the training corpus for
Church's trigram algorithm and the variable-context algorithm and both tag
sets.}

\end{figure}

However, there are also some advantages on the side of the
variable-context algorithm. First, its potential when using larger training
corpora seems to be slightly higher (see figure \ref{Results}). Secondly, when
the algorithm
is modified in such a way that sentence boundaries are not assumed to be known
beforehand, the performance degrades only minimally. This means that this
algorithm can actually contribute to finding sentence boundaries. And third,
if there are sequences of unknown word forms in the text, the algorithm takes
better guesses than the Church algorithm (examples are given in
Rapp 1995, p. 155). When about 2\% of the word forms in the test corpus were
randomly replaced by unknown word forms, the quality of the results for the
Church algorithm decreased by 0.7\% for the small and by 2.0\% for the large
tag set. The respective figures for the variable-context algorithm are 0.9\%
and 1.3\%, which is better overall.

In a further experiment, the contribution of the lexical probabilities
to the quality of the results was examined. Without the lexical
probabilities, the results decreased by 0.3\% (small) and
0.6\% (large tag set) for the Church algorithm, the respective
figures for the variable-context algorithm were 0.9\% and 0.0\%.

\begin{figure}[t]

\begin{footnotesize}
\begin{center}

\begin{tabular}{|c|c|c|c|c|c|} \hline
Die & Frau & bringt & das & Essen & . \\
\hline
ART DEF & SUB & VER & ART DEF & SUB & SZE \\ \hline
\end{tabular}

\vspace{1ex}

\begin{tabular}{|c|c|c|c|c|} \hline
Ich & meine & meine & Frau & . \\ \hline
PER PRO & VER & POS ATT & SUB &
SZE \\ \hline
\end{tabular}

\vspace{1ex}

\begin{tabular}{|c|c|c|c|c|c|c|c|c|} \hline
Winde & das & im & Winde & flatternde & Segel & um & die & Winde \\ \hline
\bf SUB & ART DEF & PRP & SUB & ADJ & SUB & PRP & ART & SUB \\ \hline
\end{tabular}

\vspace{6ex}

\begin{tabular}{|c|c|c|} \hline
Die & Frau & bringt \\ \hline
ART DEF NOM SIN FEM & SUB NOM FEM SIN & VER 3PE SIN \\ \hline
\end{tabular}

\vspace{1ex}

\begin{tabular}{|c|c|c|c|} \hline
das & Essen & . & Ich \\ \hline
ART DEF AKK SIN NEU & SUB AKK NEU SIN & SZE & PER NOM SIN 1PE \\
\hline
\end{tabular}

\vspace{1ex}

\begin{tabular}{|c|c|c|c|} \hline
meine & meine & Frau & .\\ \hline
VER 1PE SIN & POS AKK SIN FEM ATT & SUB AKK FEM SIN & SZE \\
\hline
\end{tabular}

\vspace{1ex}

\begin{tabular}{|c|c|c|} \hline
Winde & das & im \\ \hline
VER \begin{bf}3PE \end{bf} SIN & \begin{bf} DEM NOM \end{bf} SIN NEU \begin{bf}
PRO \end{bf} & PRP DAT \\ \hline
\end{tabular}

\vspace{1ex}

\begin{tabular}{|c|c|c|} \hline
Winde & flatternde & Segel \\ \hline
SUB DAT MAS SIN  & PA1 \begin{bf} SOL \end{bf} NEU AKK \begin{bf} PLU \end{bf} &
SUB AKK NEU \begin{bf} PLU \end{bf} \\ \hline
\end{tabular}

\vspace{1ex}

\begin{tabular}{|c|c|c|c|} \hline
um & die & Winde & . \\ \hline
PRP AKK & ART DEF AKK SIN FEM & SUB AKK FEM SIN & SZE \\ \hline
\end{tabular}

\end{center}

\caption{\label{Example} Tagging example for both tag sets - the
ambiguity rates amount to 2.4 tags per word for the small and 8.8 tags per word
for the large tag set (errors are printed bold type).}
\end{footnotesize}
\end{figure}


\section{Conclusions}

We have compared two different tagging algorithms and two different tag sets.
The first tagging algorithm is the Church algorithm
which uses trigrams to compute contextual probabilities. The second algorithm,
the so-called variable-context algorithm, has been described in paragraph 3.
The smaller of the two tag sets contains 51 parts-of-speech, the larger tag set
includes additional grammatical features such as case, number and gender.
The small tag set is a subset of the large tag set.

In comparison with the Church algorithm, the variable-context algorithm produces
similar results for the small tag set, but significantly inferior results for the
large tag set. On the other hand, the performance of the variable-context
algorithm for the large tag set improves faster with increasing size of the
training corpus than the performance of the Church algorithm.
Thus, with tagging more training
texts manually, similar results are to be expected for the two algorithms.

Considering the two tag sets, the results for the small tag set are
significantly better. Nevertheless, with increasing size of the training corpus
an approximation of the results seems to be possible.

One of our aims for the near future is to use the output of the tagger for
lemma\-tization. In this way a sentence like {\it Ich meine meine Frau.} could be
unambiguously reduced to {\it ich / meinen / mein / Frau}.

\vspace{-2ex}

\section*{Bibliography}

\vspace{-2ex}

S. Armstrong, G. Russell, D. Petitpierre and G. Robert (1995). An open architecture for multilingual text processing. In: {\it Proceedings of the ACL SIGDAT Workshop. From
Texts to Tags: Issues in Multilingual Language Analysis}, Dublin.

\vspace{-0.75ex}

H. Bergenholtz and B. Schaeder (1977). {\it Die
Wortarten des Deutschen}. Klett, Stuttgart.

\vspace{-0.75ex}

K. Church (1988). A stochastic parts program and noun phrase parser
for unrestricted text. In:
{\it Second Conference on Applied Natural Language Processing}, pp. 136-143.
Austin, Texas.

\vspace{-0.75ex}

D. Cutting, J. Kupiec, J. Pedersen and P. Sibun (1992). A
practical part-of-speech tagger. In: {\it Proceedings of the Third Conference on
Applied Language Processing}, pp. 133-140. Trento, Italy.

\vspace{-0.75ex}

G. Drosdowski (1984). {\it Duden. Grammatik der deutschen
Gegenwartssprache}. Dudenverlag, Mannheim.

\vspace{-0.75ex}

H. Feldweg (1995). Implementation and evaluation of a German
HMM for POS disambiguation. In: Feldweg and Hinrichs, eds., {\it Lexikon und
Text}, pp. 41-46. Niemeyer, T{\"u}bingen.

\vspace{-0.75ex}

R. Hausser (1996). {\it Linguistische Verifikation.
Dokumentation zur Ersten Morpholympics}. Niemeyer, T{\"u}bingen.

\vspace{-0.75ex}

W. Lezius (1994). Aufbau und Funktionsweise von Morphy.
Internal report. Universit{\"a}t-GH Paderborn, Fachbereich
2.

\vspace{-0.75ex}

W. Lezius (1995). Algorithmen zum Taggen deutscher Texte.
Internal report, Universit{\"a}t-GH Paderborn, Fachbereich 2.

\vspace{-0.75ex}

W. Lezius (1996). Morphologiesystem MORPHY. In:
R. Hausser, ed., {\it
Linguistische Verifikation: Dokumentation zur Ersten Morpholympics 1994},
pp. 25-35. Niemeyer, T{\"u}bingen.

\vspace{-0.75ex}

R. Rapp (1995). Die Berechnung von Assoziationen - Ein korpuslinguistischer Ansatz. In: Hellwig and Krause, eds., 
{\it Sprache und Computer}, vol. 16, Olms, Hildesheim. 
\vspace{-0.75ex}

H. Schmid (1995). Improvements in part-of-speech tagging with an
application to German. In: Feldweg and Hinrichs, eds.,
{\it Lexikon und Text}, pp. 47-50. Niemeyer, T{\"u}bingen.

\vspace{-0.75ex}

P. Steiner (1995). Anforderungen und Probleme beim Taggen
deutscher Zeitungstexte. In: Feldweg and Hinrichs, eds.,
{\it Lexikon und Text}. Niemeyer, T{\"u}bingen.

\vspace{-0.75ex}

K. Wothke, I. Weck-Ulm, J. Heinecke, O. Mertineit and T. Pachunke (1993).
Statistically Based Automatic Tagging of German Text Corpora
with Parts-of-Speech - Some Experiments. Technical Report 75.93.02,
IBM Germany, Heidelberg Scientific Center.

\end{document}